\documentclass[12pt,peerreview,draftcls]{IEEEtran}
\usepackage{mathrsfs}
\usepackage{graphicx}
\usepackage{float}
\usepackage{booktabs}
\usepackage{hyperref}
\usepackage{cite}
\usepackage{color}
\usepackage{psfrag}
\usepackage{subfigure}
\usepackage{amssymb}
\usepackage{amsthm}
\usepackage{setspace}
\usepackage{epsfig}
\usepackage{pifont}
\usepackage{amsmath}
\usepackage{array}
\usepackage{multicol}
\usepackage{multirow}
\usepackage{pifont}
\usepackage{indentfirst}
\usepackage{amsfonts}
\usepackage{algorithm}
\usepackage{algorithmicx}
\usepackage{algpseudocode}
\usepackage{fancyhdr}
\usepackage{amscd}
\usepackage{bm}
\usepackage{fancyhdr}
\usepackage{enumerate}
\usepackage{color}
\usepackage{hyperref}

\newtheorem{proposition}{Proposition}
\newtheorem{definition}{Definition}

\newcommand{\tabincell}[2]{\begin{tabular}{@{}#1@{}}#2\end{tabular}}

\IEEEoverridecommandlockouts
\begin{document}
	
	\title{Constructing Deep Neural Networks with \emph{a Priori} Knowledge of Wireless Tasks}
		
\author{
	\thanks{A part of this work is presented in conference version, which has been accepted by IEEE ICC 2020 \cite{ICC2020}.}
	\IEEEauthorblockN{Jia Guo and Chenyang Yang}\\
}
	\maketitle
	
\vspace{-4mm}\begin{abstract}
Deep neural networks (DNNs) have been employed for designing wireless systems in many aspects, say transceiver design, resource optimization, and information prediction. Existing works either use the fully-connected DNN or the DNNs with particular architectures developed in other domains. While generating labels for supervised learning and gathering training samples are time-consuming or cost-prohibitive, how to develop DNNs with wireless priors for reducing training complexity remains open. In this paper, we show that two kinds of permutation invariant properties widely existed in wireless tasks can be harnessed to reduce the number of model parameters and hence the sample and computational complexity for training. We find special architecture of DNNs whose input-output relationships satisfy the properties, called \emph{permutation invariant DNN (PINN)}, and augment the data with the properties. By learning the impact of the scale of a wireless system, the size of the constructed PINNs can flexibly adapt to the input data dimension. We take predictive resource allocation and interference coordination as examples to show how the PINNs can be employed for learning the optimal policy with unsupervised and supervised learning. Simulations results demonstrate a dramatic gain of the proposed PINNs in terms of reducing training complexity.

\begin{IEEEkeywords}
	\emph{Deep neural networks, a priori knowledge, permutation invariance, training complexity}	
\end{IEEEkeywords}
\end{abstract}

\vspace{-5mm}\section{Introduction}
Deep learning has been considered as one of the key enabling techniques in beyond fifth
generation (5G) and sixth generation (6G) cellular networks. Recently, deep neural networks (DNNs) 
have been employed to design wireless networks in various aspects, ranging from signal detection and channel estimation \cite{ye2017power,samuel2019learning}, interference management \cite{sun2017learning}, resource allocation \cite{YW19,VTC18GJ,Guo2018Exploiting,sun2019pimrc,liu2020optimizing}, coordinated beamforming \cite{alkhateeb2018deep}, traffic load prediction \cite{wang2017spatiotemporal}, and uplink/downlink channel calibration \cite{huang2019deep}, \emph{etc}, thanks to their powerful ability to learn complex input-output relation \cite{Hornik1989UnivApprox}.
For the tasks of transmission scheme or resource allocation, the output is a transceiver or allocated resource (e.g., beamforming vector or transmit power), the input is the environment parameter (e.g., channel gain), and the relation is a concerned policy (e.g.,  power allocation). For the tasks of information prediction, the relation is a predictor, which depends on the temporal correlation between historical and future samples of a time series (e.g., traffic load at a base station).

Existing research efforts focus on investigating what tasks in wireless communications can apply deep learning by  considering the fully-connected (FC)-DNN \cite{ye2017power,alkhateeb2018deep,huang2019deep,sun2017learning,sun2019pimrc}, and how deep learning is used for wireless tasks by integrating the DNNs developed in other domains such as computer vision and natural language processing \cite{wang2017spatiotemporal,VTC18GJ,YW19}. By finding the similarity between the tasks in different domains, various deep learning techniques have been employed to solve wireless problems. For example,
convolutional neural network (CNN) is applied for wireless tasks where the data exhibit spatial correlation, and recurrent neural network (RNN) is applied for information prediction using the data with temporal correlation. Most previous works consider supervised learning. Noticing the fact that generating labels is time-consuming or expensive, unsupervised learning frameworks were proposed for learning to optimize wireless systems recently \cite{YW19,liu2020optimizing}. Nonetheless, the number of samples required for training in unsupervised manner may still be very high. This impedes the practice use of DNNs  in wireless networks where data gathering is cost-prohibitive. Although the computational complexity of off-line training is less of a concern in static scenarios, wireless systems often need to operate in highly dynamic environments, where the channels, number of users, and available resources, \emph{etc.}, are time-varying. Whenever the environment parameters change, the model parameters and even the size of a DNN need to be updated (e.g., the DNN in \cite{alkhateeb2018deep} needs to be trained periodically in the timescale of minutes).  Therefore, training DNNs efficiently is critical for wireless applications.

To circumvent the ``curse of dimensionality'' that leads to the unaffordable sample and computational complexity for training, AI society has designed DNN architectures by harnessing general-purposed priors, such that each architecture is applicable for a large class of tasks. One successful example is CNN specialized for vision tasks. By exploiting the knowledge that local groups of pixels in images are often highly correlated, sparse connectivity is introduced in the form of convolution kernels. Furthermore, by exploiting the knowledge that local statistics of images are invariant to positions, parameter sharing is introduced among convolution kernels in each layer \cite{lecun2015deep, bengio2009learning}. Another example is RNN. Considering the temporal correlation feature of time series, adjacent time steps are connected with weights, and parameter sharing is introduced among time steps such that the weights between hidden layers are identical \cite{lecun2015deep}. In this way of using \emph{a priori} knowledge to design the architecture of DNNs, the number of model parameters and hence the training complexity can be reduced. To reduce the training complexity, wireless society promotes model-and-data-driven methodology to combine the well-established communication domain knowledge with deep learning most recently \cite{HJW2019,ZRD2019}. For instance, the models can be leveraged to generate labeled samples for supervised learning \cite{sun2017learning,ZRD2019}, derive gradients to guide the searching directions for stochastic gradient descent/ascent \cite{sun2019pimrc}, embed the modules with accurate models into DNN-based systems, and first use traditional model-based solutions to initialize and then apply DNNs to refine \cite{HJW2019,ZRD2019}.
Despite that the basic idea is general and useful, mathematical models are problem specific, and hence the solutions with model-based DNNs have to be developed on a case by case basis. Nonetheless, the two branches of research that are respectively priori-based and model-driven, are complementary rather than mutual exclusion.
Given the great potential of deep learning in beyond 5G/6G cellular networks, it is natural to raise the following question: are there any general priors in wireless tasks? If yes, how to design DNN architecture by incorporating the priors?

Each task corresponds to a specific relation (i.e., a function). In many wireless tasks, the relation between the concerned solutions and the relevant  parameters satisfies a common property: permutation invariance. For example, if the channel gains of multiple users permute, then the resources allocated to the users permute accordingly. This is because the resource allocated to a user depends on its own channel but not on the permutation of other users' channels \cite{sun2017learning,alkhateeb2018deep,VTC18GJ,YW19}. While the property seems obvious, the way to exploit the knowledge is not straightforward.

In this paper, we strive to demonstrate how to reduce training complexity by harnessing such \emph{general knowledge}. We consider two kinds of permutation invariance properties, which widely exist in wireless tasks. For the tasks satisfying each kind of property, we find a DNN with special architecture to represent the relation between the solution and the concerned parameters, referred to as \emph{permutation invariant DNN (PINN)}, where majority of the model parameters are identical. Different from CNN and RNN that exploit the characteristic of data, which is the input of the DNN, PINN exploits the characteristic of tasks, which decide the input-output relation.  The architecture of PINN offers the flexibility in applying to different input data dimension. By jointly trained with a small size DNN that captures the impact of the input dimension, the constructed DNNs can adapt to wireless systems with different scales (e.g., with time-varying number of users). Except the DNN architecture, we show that the property can also be used to generate labels for supervised learning.
Simulation results show that much fewer samples and much lower computational complexity are required for training the constructed PINNs to achieve a given performance, and the majority of labels can be generated with the permutation invariance property. The proposed PINNs can be applied for a broad range of wireless tasks, including but not limited to the tasks in \cite{sun2017learning,wang2017spatiotemporal,alkhateeb2018deep,VTC18GJ,YW19,huang2019deep,sun2019pimrc,samuel2019learning}.

The major contributions are summarized as follows.
\begin{itemize}
	\item We find the sufficient and necessary conditions for tasks to satisfy two kinds of permutation invariant properties. For each kind of tasks, we construct a DNN architecture whose input-output relationship satisfies the permutation invariance property. The constructed PINNs are applicable to both unsupervised and supervised learning.
	\item We show how the PINNs can adapt to different input data dimension  by introducing a factor to characterize the impact of the scale of a wireless system. In training phase, the complexities can be reduced by training DNNs with small size. In operation phase, the trained DNN can be adaptive to the input with time-varying dimension.
	\item We take predictive resource allocation and interference coordination as examples to illustrate how the PINNs can be applied to unsupervisely and supervisely learn the two kinds of permutation invariant functions, respectively. Simulation results demonstrate that the constructed PINNs can reduce the sample and computational complexities remarkably compared to the non-structural FC-DNN with same performance.
\end{itemize}

\emph{Notations}: ${\mathbb E}\{\cdot\}$ denotes mathematical expectation, $\|\cdot\|$ denotes two-norm, $\|\cdot\|_1$ denotes the summation of the absolute values of all the elements in a vector or matrix, and $(\cdot)^{\sf T}$ denotes transpose, ${\bm 1}$ denotes a column vector with all elements being $1$, ${\bm 0}$ denotes a column vector or a matrix with all elements being $0$.

The rest of the paper is organized as follows. In section \ref{sec: param share}, we introduce two permutation invariance properties and construct two PINNs, and illustrate how the PINNs can adapt to the input dimension. In section \ref{sec: case study I PRA} and \ref{sec: case study II IC}, we present two case studies. In section \ref{sec: simulation results}, we show that the PINNs can reduce training complexity, and illustrate that the properties can also be used for dataset augmentation. In section \ref{sec: conclusion}, we provide the concluding remarks.

\section{DNN for Tasks with Permutation Invariance}\label{sec: param share}
In this section, we first introduce two kinds of relationships (mathematically, two kinds of functions) with permutation invariant property, which are widely existed in wireless communication tasks. For each relationship, we demonstrate how to construct a parameter sharing DNN satisfying the property. Then, we show how to make the constructed DNN adaptive to the scale of  wireless networks.

\subsection{Definition and Example Tasks} \label{sec: definition of PI}
For many wireless tasks such as resource allocation and transceiver design, the optimized policy that yields the  solution (represented as column vector $\bf y$ without the loss of generality) from environment parameters (represented as vector $\bf x$ or matrix ${\bf X}$) can be expressed as a function ${\bf y}=f(\bf x)$ or ${\bf y}=f(\bf X)$.
Both ${\bf y}$ and ${\bf x}$ are composed of $K$ blocks, i.e., ${\bf y}=[{\bf y}_1^{\sf T}, \cdots, {\bf y}_K^{\sf T}]^{\sf T}$, ${\bf x}=[{\bf x}_1^{\sf T}, \cdots, {\bf x}_K^{\sf T}]^{\sf T}$, and ${\bf X}$ is composed of $K^2$ blocks, i.e.,
\begin{equation}\label{X}
{\bf X} = \left[
\begin{tabular}{cccc}
${\bf x}_{11}$ &  $\cdots$ & ${\bf x}_{1K}$ \\
$\vdots$ &   $\ddots$ & $\vdots$ \\
${\bf x}_{K1}$ &  $\cdots$ & ${\bf x}_{KK}$
\end{tabular}
\right],
\end{equation}
where the block ${\bf y}_k$ and ${\bf x}_k$ can either be a scalar or a column vector, $k=1,\cdots,K$, and the block ${\bf x}_{mn}$ can be a scalar, vector or matrix, $m,n=1,\cdots,K$.

A property is widely existed in the optimized policies $f(\cdot)$ for wireless problems: one-dimensional (1D) permutation invariance of ${\bf y}=f(\bf x)$ and two-dimensional (2D) permutation invariance of ${\bf y}=f(\bf X)$. Before the formal definition, we first introduce two examples.

\begin{figure}[!htb]
	\centering
	\includegraphics[width=0.6\linewidth]{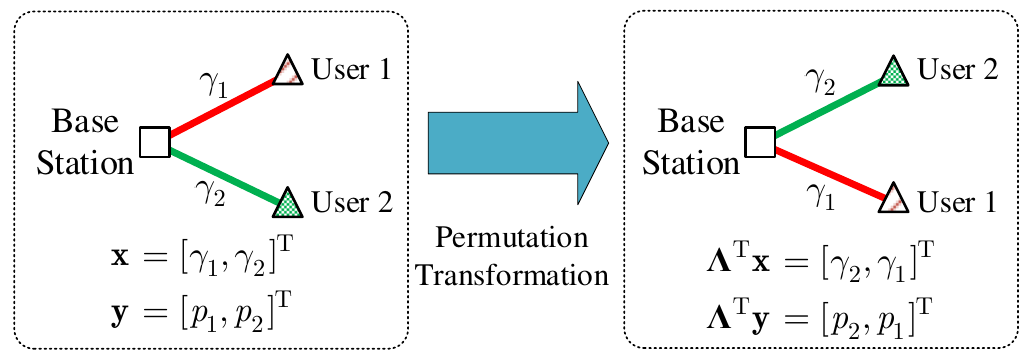}\vspace{-2mm}
	\caption{Illustration of 1D permutation invariance, power allocation to $K$ users, $K=2$.}
	\label{fig:fig-wlchnl-1d}
\end{figure}

{\bf Ex 1}: One example is the task of power allocation to $K$ users by a base station (BS), as shown in Fig. \ref{fig:fig-wlchnl-1d}. In the figure, $K=2$, each user and the BS are with a single-antenna,  $\bm \Lambda$ is a permutation matrix to be defined soon. Then, a block in ${\bf x}$, say ${\bf x}_k=\gamma_k$, is the scalar channel of the $k$th user, a block in ${\bf y}$, say ${\bf y}_k=p_k$, is transmit power allocated to the user, and ${\bf y}=f(\bf x)$ is the power allocation policy. If the users are permutated, then the allocated powers will be permutated correspondingly. Such a policy is 1D permutation invariant to ${\bf x}$.

\vspace{-1mm}\vspace{-1mm}

\begin{figure}[!htb]
	\centering
	\includegraphics[width=0.65\linewidth]{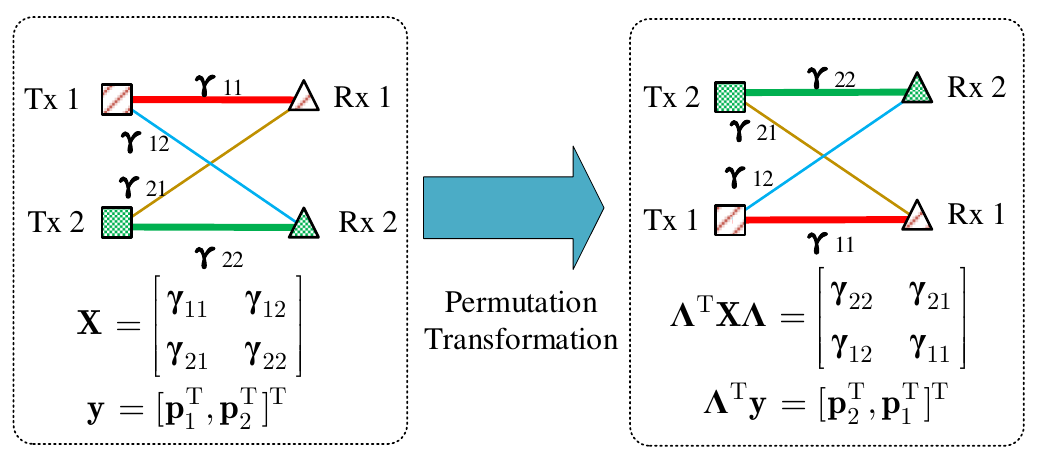}\vspace{-2mm}
	\caption{Illustration of 2D permutation invariance, interference coordination among $K$ Tx-Rx pairs, $K=2$.}
	\label{fig:fig-wlchnl}
\end{figure}

{\bf Ex 2}: Another example is the task of interference coordination among $K$ transmitters by optimizing transceivers, as shown in Fig. \ref{fig:fig-wlchnl}. Here, a block in ${\bf X}$, say ${\bf x}_{mn}={\bm\gamma}_{mn}\in {\mathbb C}^{N_{\sf tx}\times 1}$, is the channel vector between the $m$th transmitter (Tx)  and the $n$th receiver (Rx), a block in ${\bf y}$, say ${\bf y}_k={\bf p}_k\in{\mathbb R}^{N_{\sf tx}\times 1}$, is the beamforming vector for the $k$th user, $m,n,k=1,\cdots,K$, $N_{\sf tx}$ is the number of transmit antennas, and ${\bf y}=f(\bf X)$ is the interference coordination policy.
If the Tx-Rx pairs are permutated, then the beamforming vectors are correspondingly  permutated.  Such
a policy is 2D permutation invariant to ${\bf X}$.

\vspace{-2mm}\vspace{-1mm}

To define permutation invariance, we consider a column transformation matrix $\bm \Lambda$, which operates on blocks instead of the elements in each block. In other words, the permutation matrix $\bm \Lambda$ only changes the order of blocks (e.g., ${\bf x}_k$, ${\bf y}_k$ or ${\bf x}_{mn}$) but do not change the order of elements within each block (e.g., the $N_{\sf tx}$ elements in vector ${\bm \gamma}_{mn}$). An example of $\bm \Lambda$  for $K=3$ is,\vspace{-1mm}
\begin{equation}
{\bm \Lambda} = \left[
\begin{tabular}{ccc}
${\bf I}$ & ${\bf 0}$  & ${\bf 0}$ \\
${\bf 0}$ & ${\bf 0}$  & ${\bf I}$ \\
${\bf 0}$ & ${\bf I}$ & ${\bf 0}$
\end{tabular}
\right], \notag
\end{equation}\vspace{-1mm}
where ${\bf I}$ and ${\bf 0}$ are respectively the identity matrix and square matrix with all zeros.

\vspace{-1mm}\begin{definition} \label{def: 1}
	For arbitrary permutation to ${\bf x}$, i.e., ${\bm \Lambda}^{\sf T}{\bf x}=[{\bf x}_{N_1}^{\sf T},\cdots,{\bf x}_{N_K}^{\sf T}]^{\sf T}$ where $N_1,\cdots,N_K$ is arbitrary permutation of $1,\cdots,K$, if ${\bm \Lambda}^{\sf T}{\bf y}=f({\bm \Lambda}^{\sf T}{\bf x})=[{\bf y}_{N_1}^{\sf T},\cdots,{\bf y}_{N_K}^{\sf T}]^{\sf T}$, then $f({\bf x})$ is 1D permutation invariant to ${\bf x}$.
\end{definition}\vspace{-1mm}

In the following, we provide the sufficient and necessary condition for a function ${\bf y}=f(\bf x)$ to be 1D permutation invariant.
\vspace{-1mm}\begin{proposition}\label{pp: 1}
	The function ${\bf y}=f({\bf x})$ is 1D permutation invariant to ${\bf x}$ if and only if,
	\begin{equation} \label{eq: perm inva}
	{\bf y}_k = \eta\Big(\psi({\bf x}_k), {\cal F}_{n=1,n\neq k}^K \phi({\bf x}_n)\Big), k=1,\cdots,K,
	\end{equation}
	where $\eta(\cdot), \psi(\cdot)$ and $\phi(\cdot)$ are arbitrary functions, and ${\cal F}$ is arbitrary operation satisfying the commutative law.
	\begin{IEEEproof}
		See Appendix \ref{appendix: A}.	
	\end{IEEEproof}
\end{proposition}\vspace{-3mm}

The operations satisfying the commutative law include summation, product, maximization and minimization, \emph{etc}. To help understand this condition, consider a more specific class of functions ${\bf y}=f(\bf x)$ satisfying \eqref{eq: perm inva}, where the $k$th block in ${\bf y}$ can be expressed as
\begin{equation} \label{eq: perm inva-ex}
{\bf y}_k = \eta\Big(\psi({\bf x}_k), \sum_{n=1,n\neq k}^K \phi({\bf x}_n)\Big), k=1,\cdots,K.
\end{equation}
Such class of functions ${\bf y}=f({\bf x})$ are 1D permutation invariant to ${\bf x}$. This is because for any permutation of ${\bf x}$, $\tilde{\bf x}=[{\bf x}_{N_1}^{\sf T},\cdots, {\bf x}_{N_K}^{\sf T}]^{\sf T}$, the solution corresponding to $\tilde{\bf x}$ is $\tilde{\bf y}=[\tilde{\bf y}_1^{\sf T},\cdots,\tilde{\bf y}_K^{\sf T}]^{\sf T}=[{\bf y}_{N_1}^{\sf T},\cdots,{\bf y}_{N_K}^{\sf T}]^{\sf T}$, where the $k$th block of $\tilde{\bf y}$ is $\tilde{\bf y}_k = \eta(\psi({\bf x}_{N_k}),\sum_{n=1,n\neq N_k}^K \phi({\bf x}_n))={\bf y}_{N_k}$.

For {\bf Ex 1}, the optimal power allocation can be expressed as \eqref{eq: perm inva-ex} (though may not be explicitly),  where $\psi({\bf x}_k)$ reflects the impact of the $k$th user's channel on its own power allocation, and $\sum_{n=1,n\neq k}^K \phi({\bf x}_n)$ reflects the impact of other users' channels on the power allocation to the $k$th user.
From \eqref{eq: perm inva} or \eqref{eq: perm inva-ex} we can observe that: (i) the impact of the block ${\bf x}_k$ and the impact of other blocks ${\bf x}_n,n\neq k$ on ${\bf y}_k$ are different, and (ii) the impact of every single block ${\bf x}_n, n\neq k$ on ${\bf y}_k$ does not need to be differentiated.

This suggests that for a DNN to learn the permutation invariant functions, it should and only need to compose of two types of weights to respectively reflect the two kinds of impact.

\vspace{-2mm}\begin{definition} \label{def: 2}
	For arbitrary permutation to the columns and rows of ${\bf X}$, i.e., ${\bm \Lambda}^{\sf T}{\bf X}{\bm \Lambda}$, if ${\bm \Lambda}^{\sf T}{\bf y}=f({\bm \Lambda}^{\sf T}{\bf X}{\bm \Lambda})$, then $f({\bf X})$ is 2D permutation invariant to ${\bf X}$.
\end{definition}\vspace{-2mm}

Using the similar method as in Appendix \ref{appendix: A}, we can prove the following sufficient and necessary condition for a function ${\bf y}=f(\bf X)$ to be 2D permutation invariant.
\vspace{-2mm}\begin{proposition}\label{pp: 3}
	The function ${\bf y}=f({\bf X})$ is permutation invariant to ${\bf X}$  if and only if,
	\begin{equation} \label{eq: perm inva2}
	{\bf y}_k = \eta\Big(\psi({\bf x}_{kk}), {\cal F}_{n=1,n\neq k}^K \phi({\bf x}_{kn}), {\cal G}_{n=1,n\neq k}^K \xi({\bf x}_{nk}), {\cal H}_{m,n=1,m,n\neq k}^K \zeta({\bf x}_{mn})\Big), k=1,\cdots,K,
	\end{equation}
	where $\eta(\cdot), \zeta(\cdot), \psi(\cdot), \xi(\cdot)$ and $\phi(\cdot)$ are arbitrary functions, and ${\cal F}, {\cal G}, {\cal H}$ are arbitrary operations satisfying the commutative law.
\end{proposition}\vspace{-2mm}

Similarly, from \eqref{eq: perm inva2} we can observe that: (i) the impact of ${\bf x}_{kk}$, $\{{\bf x}_{kn},n\neq k\}$, $\{{\bf x}_{nk},n\neq k\}$, $\{{\bf x}_{mn},m,n\neq k\}$ on ${\bf y}_k$ are different, (ii) the impact of every single block ${\bf x}_{kn},n\neq k$ (also ${\bf x}_{nk},n\neq k$ and ${\bf x}_{mn},m,n\neq k$) on ${\bf y}_k$ does not need to be differentiated.

For {\bf Ex 2}, the optimized solution for a Tx-Rx pair (say ${\bf p}_1$ for the first pair) depends on the channels of four links: (i) the channel between Tx1 and Rx1 ${\bm\gamma}_{11}$, (ii) the channels between Tx1  and other receivers ${\bm\gamma}_{1k},k=1,\cdots,K, k\neq 1$, (iii) the channels between other transmitters and Rx1  ${\bm\gamma}_{k1},k=1,\cdots,K, k\neq 1$, and (iv) the channels between all the other transmitters and receivers ${\bm\gamma}_{mn},m,n=1,\cdots,K, m,n\neq 1$. Their impacts on ${\bf y}_k$ are reflected respectively by the four terms within the outer bracket of \eqref{eq: perm inva2}.   When $N_{\sf tx}=1$, both ${\bm\gamma}_{mn}$ and ${\bf p}_k$ become scalers, while ${\bf y}=f({\bf X})$ is still 2D permutation invariant to ${\bf X}$.

\vspace{-5mm}\subsection{DNN Architectures for the Tasks with One- and Two-dimensional Permutation Invariance} \label{sec: 1d-perm inva}
When we design DNNs for wireless tasks such as resource allocation, the essential goal of a DNN is to learn a function
${\bf y}=f({\bf x}, {\bf W})$ or ${\bf y}=f({\bf X}, {\bf W})$ ,
where ${\bf x}$ or ${\bf X}$ and ${\bf y}$ are respectively the input and output of the DNN, and ${\bf W}$ is the model parameters that need to be trained.

In what follows, we demonstrate how to construct the  architecture of the DNN for the tasks whose policies have the property of 1D or 2D permutation invariance.
\subsubsection{One-dimensional Permutation Invariance}
To begin with, consider the FC-DNN, which has no particular architecture and hence can approximate arbitrary function. The input-output relation of a FC-DNN consisting of $L$ layers can be expressed as,
\vspace{-2mm}
\begin{equation}\label{eq: FC func}
{\bf y}\!=\!f({\bf x}, {\bf W})\!\triangleq\! g^{[L]}\left({\bf W}^{[L-1,L]} g^{[L-1]}\Big(\cdots g^{[2]}({\bf W}^{[1,2]}{\bf x}+{\bf b}^{[2]})\cdots\Big)+{\bf b}^{[L]}\right),
\end{equation}
where ${\bf W} = \{\{{\bf W}^{[l-1,l]}\}_{l=2}^L, \{{\bf b}^{[l]}\}_{l=2}^L\}$ represents the model parameters.

When $f({\bf x}, {\bf W})$ is 1D permutation invariant to ${\bf x}$, we can reduce the number of model parameters by introducing parameter sharing among the blocks  into the FC-DNN.
Inspired by the observation from \eqref{eq: perm inva} or \eqref{eq: perm inva-ex}, we can construct a DNN with a special architecture to learn a 1D permutation invariant function. Denote the output of the $l$th hidden layer as ${\bf h}^{[l]}$. Then, the relation between ${\bf h}^{[l]}$ and ${\bf h}^{[l-1]}$ is,\vspace{-1mm}
\begin{equation}
{\bf h}^{[l]} = g^{[l]}({\bf W}^{[l-1,l]}{\bf h}^{[l-1]}+{\bf b}^{[l]}),
\end{equation}
with the weight matrix between the $(l-1)$th layer and the $l$th layer as,\vspace{-1mm}
\begin{equation}\label{weight mat0}
{\bf W}^{[l-1,l]} = \left[
\begin{tabular}{cccc}
${\bf U}^{[l-1,l]}$ & ${\bf V}^{[l-1,l]}$ & $\cdots$ & ${\bf V}^{[l-1,l]}$ \\
${\bf V}^{[l-1,l]}$ & ${\bf U}^{[l-1,l]}$ & $\cdots$ & ${\bf V}^{[l-1,l]}$ \\
$\vdots$ & $\vdots$ & $\ddots$ & $\vdots$ \\
${\bf V}^{[l-1,l]}$ & ${\bf V}^{[l-1,l]}$ & $\cdots$ & ${\bf U}^{[l-1,l]}$
\end{tabular}
\right],
\end{equation}
where ${\bf U}^{[l-1,l]}$ and ${\bf V}^{[l-1,l]}$ are sub-matrices with the numbers of rows and columns respectively equal to the numbers of elements in ${\bf h}^{[l]}_k$ and ${\bf h}^{[l-1]}_k$, and ${\bf h}^{[l]}_k$ and ${\bf h}^{[l-1]}_k$ are respectively the $k$th block in the output of the $l$th and $(l-1)$th hidden layers, $k=1,\cdots,K, l=2,\cdots,L$. ${\bf b}^{[l]}=[({\bf a}^{[l]})^{\sf T},\cdots,({\bf a}^{[l]})^{\sf T}]^{\sf T}$, ${\bf a}^{[l]}$ is sub-vector with number of elements equal to that of ${\bf h}^{[l]}_k$. $g^{[l]}(\cdot)$ is the element-wise activation function of the $l$th layer.

When $l=1, {\bf h}^{[l]}={\bf x}$, and when $l=L, {\bf h}^{[l]}={\bf y}$.

\vspace{-2mm}\begin{proposition} \label{pp: 2}
When the weight matrices ${\bf W}^{[l-1,l]},l=2,\cdots,L$ are with the structure in \eqref{weight mat0}, ${\bf y}=f({\bf x}, {\bf W})$ in \eqref{eq: FC func} is 1D permutation invariant to ${\bf x}$.
\begin{IEEEproof}
	 For notational simplicity, we omit the bias vector in this proof. With the weight matrices in \eqref{weight mat0}, the output of the 2nd hidden layer is ${\bf h}^{[2]} = g^{[2]}({\bf W}^{[1,2]}{\bf x})$, and the output of the $l$th hidden layer $(2<l<L)$ can be written as,\vspace{-2mm}
	\begin{eqnarray}\label{B1}
	&{\bf h}^{[l]}& \notag\\ &=& g^{[l]}({\bf W}^{[l-1,l]}{\bf h}^{[l-1]}) \notag\\
	&=&\Big[g^{[l]}\big({\bf U}^{[l-1,l]}{\bf h}^{[l-1]}_1+{\bf V}^{[l-1,l]}\sum_{k=2}^K {\bf h}^{[l-1]}_k\big),\cdots,\notag  g^{[l]}\big({\bf U}^{[l-1,l]}{\bf h}^{[l-1]}_K+{\bf V}^{[l-1,l]}\sum_{k=1}^{K-1} {\bf h}^{[l-1]}_k\big)\Big],
	\end{eqnarray}
	where ${\bf h}^{[l]}=[({\bf h}^{[l]}_1)^{\sf T},\cdots,({\bf h}^{[l]}_K)^{\sf T}]^{\sf T}$. The $k$th block of ${\bf h}^{[l]}$ can be expressed as
	\begin{equation}\label{eq: hlk}
	{\bf h}^{[l]}_k=g^{[l]}\left({\bf U}^{[l-1,l]}{\bf h}^{[l-1]}_k\!+\!{\bf V}^{[l-1,l]}\sum_{n=1,n\neq k}^K {\bf h}^{[l-1]}_n\right).
	\end{equation}
	
	We can see that the relation between ${\bf h}^{[l]}_k$ and ${\bf h}^{[l-1]}_k$ has the same form as in \eqref{eq: perm inva}. Then, according to Proposition 1, ${\bf h}^{[l]}=g^{[l]}({\bf W}^{[l-1,l]}{\bf h}^{[l-1]})$ is 1D permutation invariant to ${\bf h}^{[l-1]}$.
	
	Since the output of every hidden layer is permutation invariant to the output of its previous layer, and ${\bf y} =g^{[L]}({\bf W}^{[L-1,L]}{\bf h}^{[L-1]})$ is also permutation invariant to ${\bf h}^{[L-1]}$, $f({\bf x}, {\bf W})$ in \eqref{eq: FC func} is 1D permutation invariant to ${\bf x}$.
\end{IEEEproof}
\end{proposition}\vspace{-2mm}

To help understand how a function with 1D permutation invariance property is constructed by the DNN in Proposition 3, consider a neural network with no hidden layers, and omit the superscript $[l-1,l]$ and the bias for easy understanding. Then, the $k$th output of the neural network can be expressed as $\textstyle {\bf y}_k = g({\bf U}{\bf x}_k + {\sum_{n=1,n\neq k}^K \bf V} {\bf x}_n)$. By comparing with \eqref{eq: perm inva-ex}, we can see that $\eta(\cdot)$ is constructed as the activation function $g(\cdot)$, $\psi(\cdot)$  and $\phi(\cdot)$ are respectively constructed as linear functions as $\psi({\bf x}_k)={\bf U}{\bf x}_k$ and  $\phi({\bf x}_k)={\bf V}{\bf x}_k$, and the operation $\cal F$ is $\sum_{n=1,n\neq k}$.

\begin{figure}[!htb]
	\centering
	\vspace{-3mm}
	\includegraphics[width=.5\linewidth]{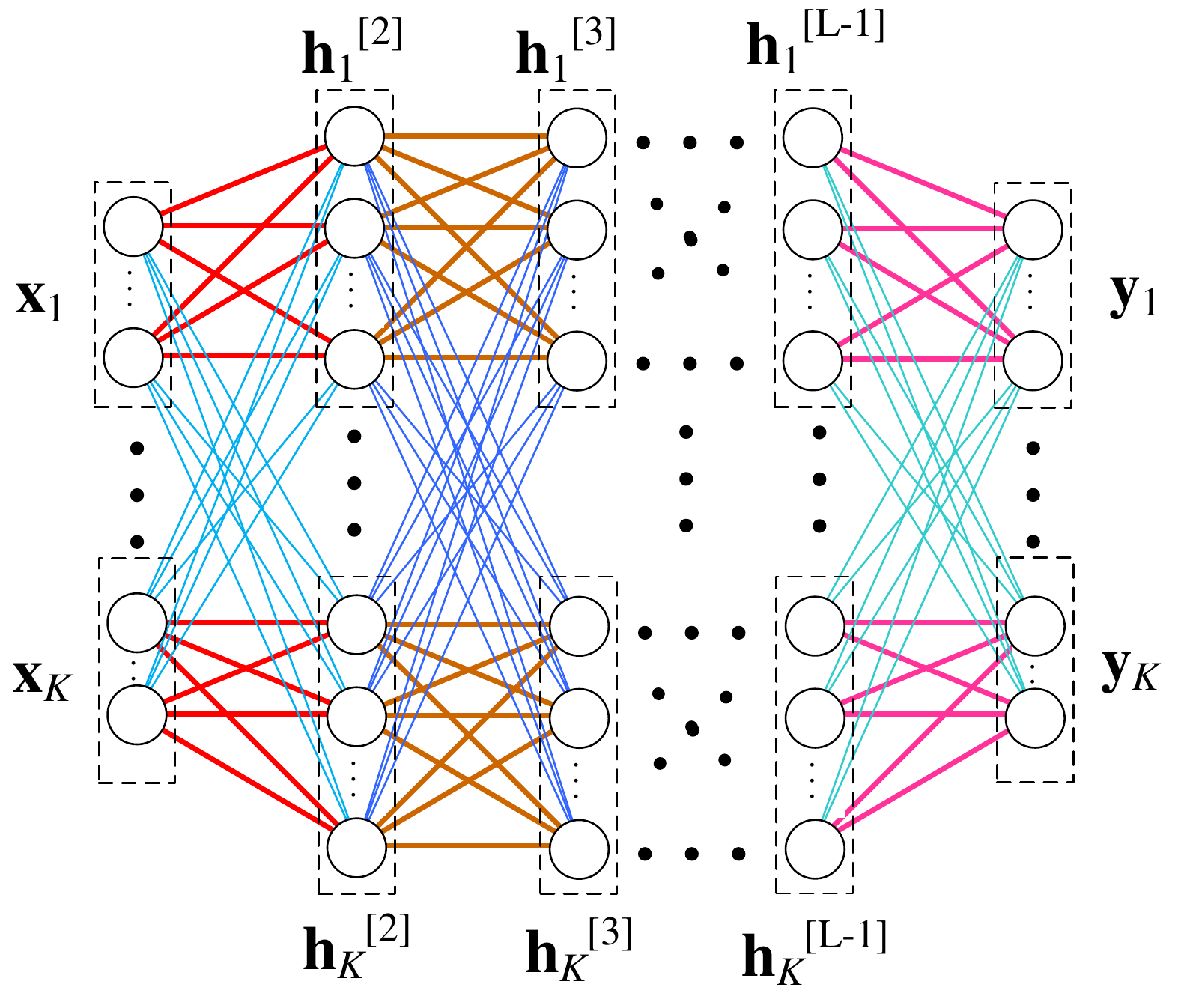}
	\vspace{-2mm}
	\caption{Architecture of PINN-1D. The connections with the same color are with same sub-weight matrices (i.e., ${\bf U}^{[l-1,l]}$ and ${\bf V}^{[l-1,l]}$). The neurons within the dashed box belong to a block. ${\bf h}^{[l]}_k$ denotes the $k$th block in the output of the $l$th hidden layer.}
	\vspace{-2mm}
	\label{fig:dnn}
	\vspace{-2mm}
\end{figure}

In \eqref{weight mat0}, all the diagonal sub-matrices of ${\bf W}^{[l-1,l]}$ are ${\bf U}^{[l-1,l]}$, which are the model parameters to learn the impact of ${\bf x}_k$ on ${\bf y}_k$. All the other sub-matrices are ${\bf V}^{[l-1,l]}$, which are the model parameters to learn the impact of ${\bf x}_n, n \neq k$ on ${\bf y}_k$. Since only two (rather than $K^2$ as in FC-DNN) sub-matrices need to
be trained in each layer, the training complexity of the DNN can be reduced. We refer this DNN with 1D permutation invariance property as ``\textbf{PINN-1D}'', which shares parameters among blocks in each layer as shown in Fig. \ref{fig:dnn}.
.

\subsubsection{Two-dimensional Permutation Invariance} \label{sec: 2d-perm inva}
\begin{figure}[!htb]
	\centering
	\includegraphics[width=.95\linewidth]{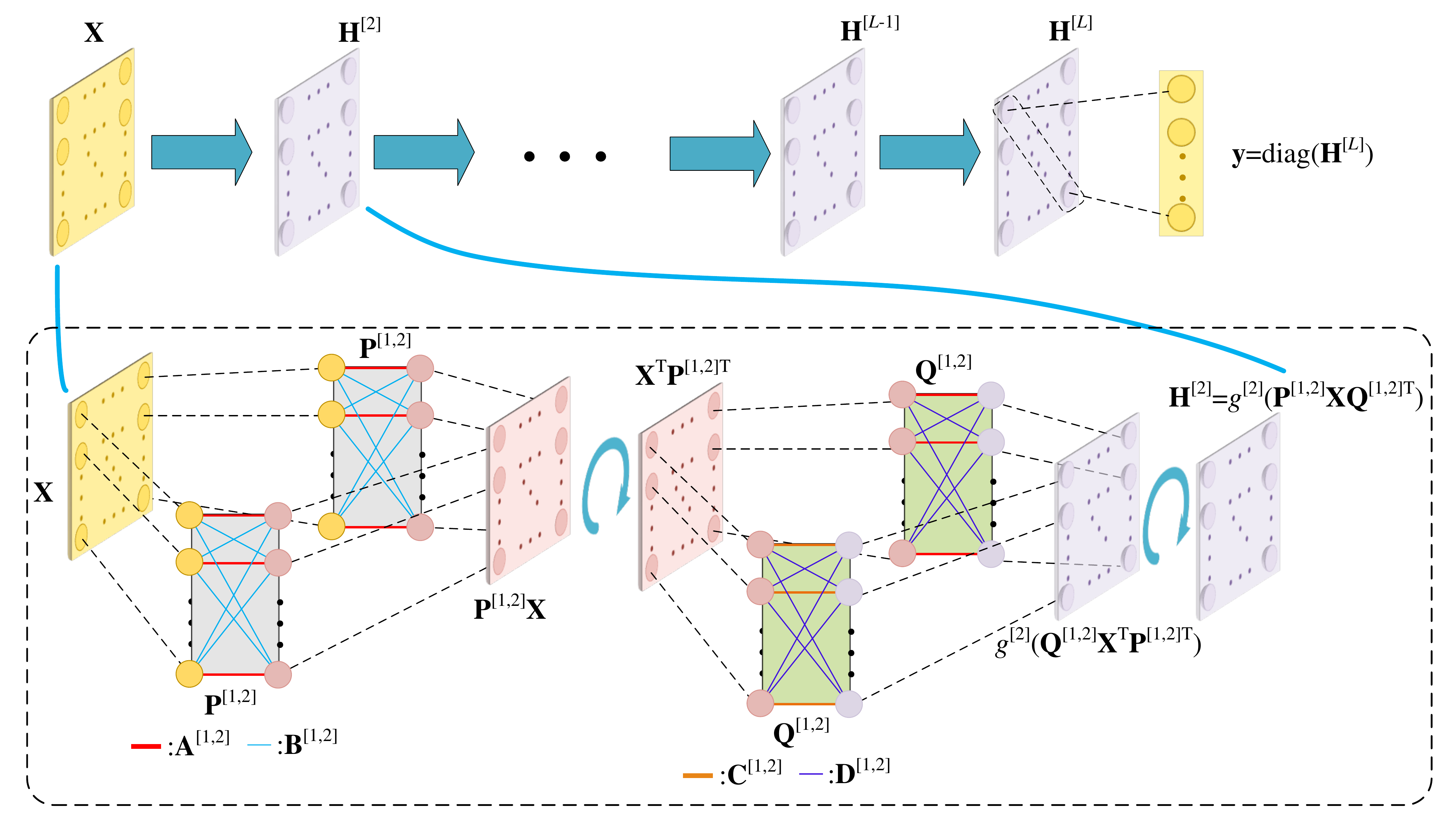}\vspace{-2mm}
	\caption{Architecture of PINN-2D. Each circle represents a block instead of a neuron.}
	\label{fig:dnn-2d}
\end{figure}

When $f({\bf X}, {\bf W})$ is 2D permutation invariant to ${\bf X}$, we can also reduce the number of model parameters by sharing parameters among blocks, as inspired by the observation from \eqref{eq: perm inva2}. The constructed ``\textbf{PINN-2D}'' is shown in Fig. \ref{fig:dnn-2d}.

\vspace{-2mm}

Different from ``\textbf{PINN-1D}'', the output of each layer is a matrix instead of a vector. Denote the output of the $l$th hidden layer as ${\bf H}^{[l]}$.  To learn a 2D permutation invariant function, the relation between ${\bf H}^{[l]}$ and ${\bf H}^{[l-1]}$ is constructed as,
\begin{equation}\label{eq: 2DPS-hidden}
{\bf H}^{[l]} = g^{[l]}\Big({\bf P}^{[l-1,l]}{\bf H}^{[l-1]} ({\bf Q}^{[l-1,l]})^{\sf T}\Big),
\end{equation}
with the weight matrices between the $(l-1)$th layer and the $l$th layer as,
\begin{equation}\label{weight mat1}
{\bf P}^{[l-1,l]} \!=\!\! \left[\!\!
\begin{tabular}{cccc}
${\bf A}^{[l-1,l]}$ & ${\bf B}^{[l-1,l]}$ & $\cdots$ & ${\bf B}^{[l-1,l]}$ \\
${\bf B}^{[l-1,l]}$ & ${\bf A}^{[l-1,l]}$ & $\cdots$ & ${\bf B}^{[l-1,l]}$ \\
$\vdots$ & $\vdots$ & $\ddots$ & $\vdots$ \\
${\bf B}^{[l-1,l]}$ & ${\bf B}^{[l-1,l]}$ & $\cdots$ & ${\bf A}^{[l-1,l]}$
\end{tabular}
\!\!\right]\!\!,
{\bf Q}^{[l-1,l]} \!=\!\! \left[\!\!
\begin{tabular}{cccc}
${\bf C}^{[l-1,l]}$ & ${\bf D}^{[l-1,l]}$ & $\cdots$ & ${\bf D}^{[l-1,l]}$ \\
${\bf D}^{[l-1,l]}$ & ${\bf C}^{[l-1,l]}$ & $\cdots$ & ${\bf D}^{[l-1,l]}$ \\
$\vdots$ & $\vdots$ & $\ddots$ & $\vdots$ \\
${\bf D}^{[l-1,l]}$ & ${\bf D}^{[l-1,l]}$ & $\cdots$ & ${\bf C}^{[l-1,l]}$
\end{tabular}
\!\!\right]\!\!,
\end{equation}
where ${\bf A}^{[l-1,l]}$ and ${\bf B}^{[l-1,l]}$ are sub-matrices with the number of rows and columns respectively equal to the number of rows in ${\bf h}^{[l]}_{mn}$ and in ${\bf h}^{[l-1]}_{mn}$, ${\bf C}^{[l-1,l]}$ and ${\bf D}^{[l-1,l]}$ are sub-matrices with the number of rows and columns respectively equal to the number of columns in ${\bf h}^{[l]}_{mn}$ and in ${\bf h}^{[l-1]}_{mn}$, ${\bf h}^{[l]}_{mn}$ is the block in the $m$th row of the $n$th column of ${\bf H}^{[l]}$, $g^{[l]}$ is the element-wise activation function of the $l$th layer, $m,n=1,\cdots,K, l=2,\cdots,L$.

We can see from \eqref{weight mat1} that both ${\bf P}^{[l-1,l]}$ and ${\bf Q}^{[l-1,l]}$ consist of two sub-matrices, where one of them is on the diagonal position and the other one is on the off-diagonal position.

Since the output of the DNN is a vector while the output of the last hidden layer ${\bf H}^{[L]}$ is a matrix, to satisfy permutation invariance we let ${\bf y}={\cal E}({\bf H}^{[L]})$  in the last layer, where ${\cal E}(\cdot)$ can be arbitrary operation satisfying ${\bf \Lambda}^{\sf T}{\bf y}={\cal E}({\bf \Lambda}^{\sf T}{\bf H}^{[L]}{\bf \Lambda})$. As an illustration, we set ${\bf y}$ as the diagonal elements of ${\bf H}^{[L]}$, i.e., ${\bf y}_k={\bf h}^{[L]}_{kk}, k=1,\cdots,K$. Then, the input-output relation of the constructed PINN-2D can be expressed as,
\begin{equation}\label{eq: PINN func}
{\bf y}=f({\bf X}, {\bf W})\triangleq {\sf diag}\left(g^{[L]}\Big({\bf P}^{[L-1,L]} g^{[L-1]}\big(\cdots g^{[2]}({\bf P}^{[1,2]}{\bf X}({\bf Q}^{[1,2]})^{\sf T}\cdots\big)({\bf Q}^{[L-1,L]})^{\sf T}\Big)\right),
\end{equation}
where ${\bf W}=\{{\bf A}^{[l-1,l]}, {\bf B}^{[l-1,l]}, {\bf C}^{[l-1,l]}, {\bf D}^{[l-1,l]}\}_{l=2}^L$, and ${\sf diag}(\cdot)$ denotes the operation of concatenating diagonal blocks of a matrix into a vector.

\vspace{-2mm}\begin{proposition}
	When the weight matrices ${\bf P}^{[l-1,l]}$ and ${\bf Q}^{[l-1,l]}$ are with the structure in \eqref{weight mat1}, ${\bf y}=f({\bf X}, {\bf W})$ in \eqref{eq: PINN func} is 2D permutation invariant to ${\bf X}$.
	\begin{proof}
		With ${\bf P}^{[l-1,l]}$ and ${\bf Q}^{[l-1,l]}$ in \eqref{weight mat1}, for arbitrary column transformation ${\bm \Lambda}$, it is easy to prove that ${\bf P}^{[l-1,l]}{\bm \Lambda}^{\sf T}={\bm \Lambda}^{\sf T} {\bf P}^{[l-1,l]}$ and ${\bf Q}^{[l-1,l]}{\bm \Lambda}^{\sf T}={\bm \Lambda}^{\sf T} {\bf Q}^{[l-1,l]}, l=2,\cdots,L$. Since $g^{[l]},l=2,\cdots,L$ are element-wise activation functions, from \eqref{eq: 2DPS-hidden} we have
		\begin{eqnarray}
		{\bm \Lambda}^{\sf T}{\bf H}^{[l]}{\bm \Lambda}
		&=& {\bm \Lambda}^{\sf T}g^{[l]}\Big({\bf P}^{[l-1,l]}{\bf H}^{[l-1]} ({\bf Q}^{[l-1,l]})^{\sf T}\Big){\bm \Lambda}
		= g^{[l]}\Big({\bm \Lambda}^{\sf T}{\bf P}^{[l-1,l]}{\bf H}^{[l-1]} ({\bf Q}^{[l-1,l]})^{\sf T}{\bm \Lambda}\Big) \notag\\
		&=& g^{[l]}\Big({\bf P}^{[l-1,l]}{\bm \Lambda}^{\sf T}{\bf H}^{[l-1]}{\bm \Lambda} ({\bf Q}^{[l-1,l]})^{\sf T}\Big). \notag
		\end{eqnarray}
Further considering that it is easy to prove that ${\bm \Lambda}^{\sf T}{\bf y} = {\sf diag}({\bm \Lambda}^{\sf T}{\bf H}^{[L]}{\bm \Lambda})$, from \eqref{eq: PINN func} we have ${\bm \Lambda}^{\sf T}{\bf y}=f({\bm \Lambda}^{\sf T}{\bf X}{\bm \Lambda}, {\bf W})$. Then, according to Definition \ref{def: 2} we know that ${\bf y}=f({\bf X}, {\bf W})$ is 2D permutation invariant to ${\bf X}$.
	\end{proof}
\end{proposition}\vspace{-2mm}

To show how a function with 2D permutation invariance property is constructed by such a DNN, consider a  PINN-2D with one hidden layer and omit the subscript $[l-1, l]$ and $[l]$ for notational simplicity. Then, relation between ${\bf y}_k$ and ${\bf X}$ can be obtained from \eqref{weight mat1} and \eqref{eq: PINN func} as,
\begin{equation} \label{eq: 2DPS-xy}
{\bf y}_k=g\left({\bf A}{\bf x}_{kk}{\bf C}^{\sf T}
+\sum_{n=1,n\neq k}^K {\bf A}{\bf x}_{kn}{\bf D}^{\sf T}
+\sum_{n=1,n\neq k}^K {\bf B}{\bf x}_{nk}{\bf C}^{\sf T}
+\sum_{m,n=1,m,n\neq k}^K {\bf B}{\bf x}_{mn}{\bf D}^{\sf T}\right),
\end{equation}
which has the same form as \eqref{eq: perm inva2}, and the sub-matrices ${\bf A}, {\bf B}, {\bf C}, {\bf D}$ are used to learn the impact of ${\bf x}_{kk}, \{{\bf x}_{kn}, n\neq k\}, \{{\bf x}_{nk}, n\neq k\}, \{{\bf x}_{mn}, m,n\neq k\}$ on ${\bf y}_k$. By comparing \eqref{eq: perm inva2} and \eqref{eq: 2DPS-xy},
we can see that $\eta(\cdot)$ is constructed as the activation function $g(\cdot)$, the functions $\psi(\cdot)$,  $\phi(\cdot)$, $\xi(\cdot)$ and $\zeta(\cdot)$ are respectively constructed as bi-linear functions as $\psi({\bf x}_{kk})={\bf A}{\bf x}_{kk}{\bf C}^{\sf T}$,
$\phi({\bf x}_{kn})={\bf A}{\bf x}_{kn}{\bf D}^{\sf T}$,
$ \xi({\bf x}_{nk})=  {\bf B}{\bf x}_{nk}{\bf C}^{\sf T}$,
and $\zeta({\bf x}_{mn})= {\bf B}{\bf x}_{mn}{\bf D}^{\sf T}$. The operations ${\cal F}$ and ${\cal G}$ are $\sum_{n=1,n\neq k}^K$, and the operation ${\cal H}$ is $\sum_{m,n=1,m,n\neq k}^K$.

In the sequel, and refer both PINN-1D and PINN-2D as ``\textbf{PINN}'' when we do not need to differentiate them.

\vspace{-2mm}\subsection{Network Size Adaptation} \label{sec: net size adapt}
The PINN is organized in blocks, i.e., the numbers of blocks in the input, output and hidden layers depend on $K$. In practice, the value of $K$, e.g., the number of users in a cell, is time-varying.
In the following, we take PINN-1D as an example to illustrate how to make PINN adaptive to $K$, while PINN-2D can be designed in the same way.

\begin{figure}[!htb] \vspace{-2mm}
	\centering
	\includegraphics[width=0.7\linewidth]{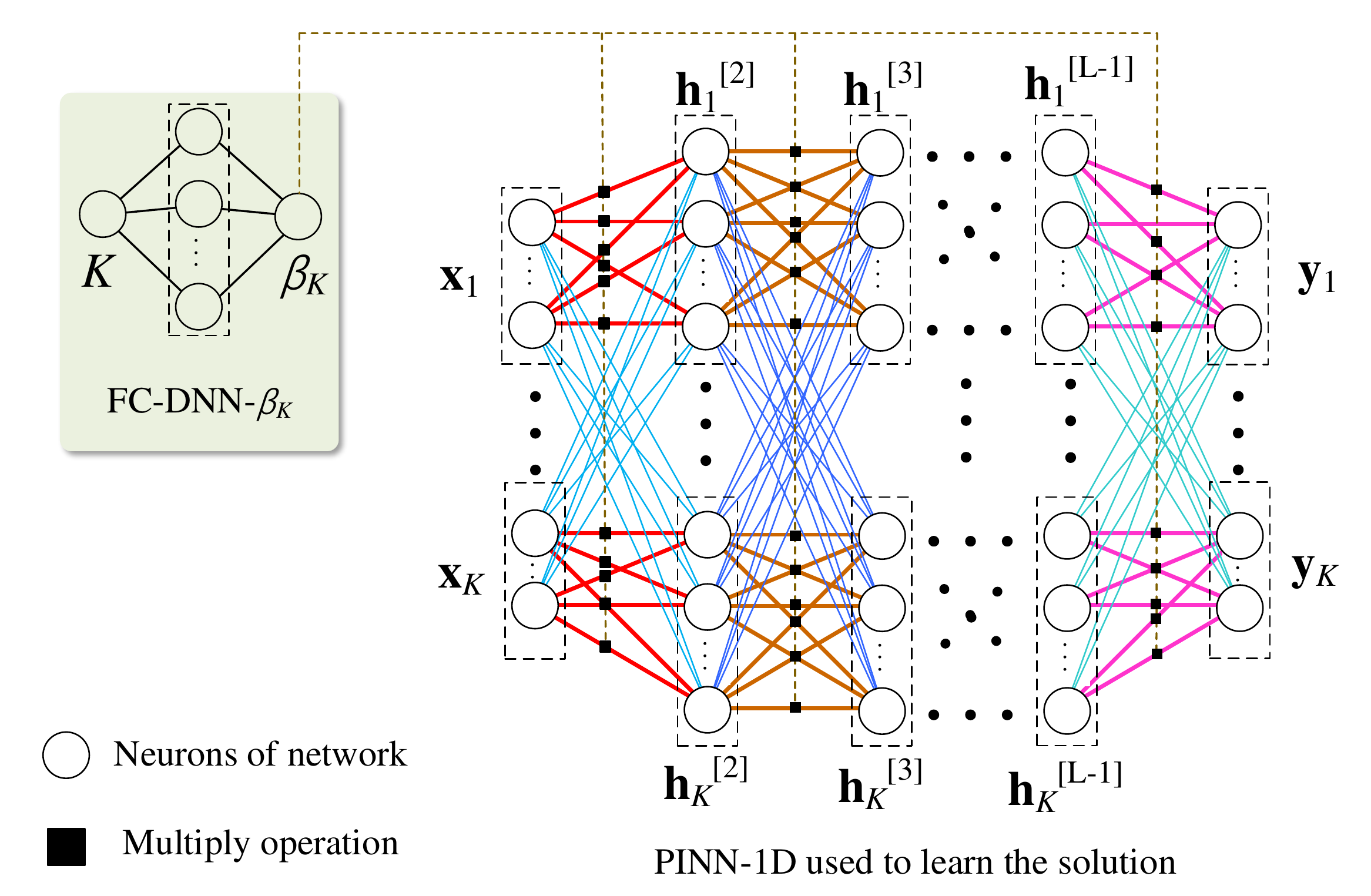}\vspace{-2mm}
	\caption{Illustration of the PINN-1D architecture that can adapt to $K$, referred to as PINN-1D-Adp-$K$.}
	\label{fig:dnn-dimgen}
\end{figure}

In each layer (say the $l$th layer) of PINN-1D, the matrix ${\bf W}^{[l-1,l]}$ with $K^2$ blocks is composed of two sub-matrices ${\bf U}^{[l-1,l]}$ and ${\bf V}^{[l-1,l]}$, where each block corresponds to one of the sub-matrices.
Therefore, the size of ${\bf W}^{[l-1,l]}$ can be flexibly controlled by adding or removing sub-matrices to adapt to different values of $K$.
It is shown from \eqref{eq: hlk} that the impact of other blocks in the $(l-1)$th hidden layer on ${\bf h}^{[l]}_k$ grows with the value of $K$. When $K$ is large, the impact of ${\bf h}^{[l-1]}_k$ on ${\bf h}^{[l]}_k$ (i.e., the first term in  \eqref{eq: hlk}) diminishes. To avoid this,
we multiply the sub-matrix ${\bf U}^{[l-1,l]}$ with a factor $\beta_K$ that is learned by a FC-DNN (denoted as FC-DNN-$\beta_K$) with the input as $K$, as shown in Fig. \ref{fig:dnn-dimgen}. Then, the $k$th block of the output of the $l$th hidden layer becomes,
\begin{equation}\label{eq: dimgen}
{\bf h}^{[l]}_k=g^{[l]}\left(\beta_K{\bf U}^{[l-1,l]}{\bf h}^{[l-1]}_k+{\bf V}^{[l-1,l]}\sum_{n=1,n\neq k}^K {\bf h}^{[l-1]}_n\right).
\end{equation}
In this way, the DNN can adaptive to $K$. We call the DNN in Fig. \ref{fig:dnn-dimgen} as ``\textbf{PINN-1D-Adp-$K$}''.
 \vspace{-0.1mm}

The model parameters of PINN-1D and FC-DNN-$\beta_K$ are jointly trained. Specifically, since $\beta_K$ is learned by inputting $K$, the relation between $\beta_K$ and $K$ can be written as  $\beta_K=f_{\beta}(K, {\bf W}_{\beta})$, where ${\bf W}_{\beta}$ is the model parameters in FC-DNN-$\beta_K$. Then, the relation between output $\bf y$ and input ${\bf x}$ can be written as ${\bf y}=f({\bf x}, {\bf W}, f_{\beta}(K, {\bf W}_{\beta}))$, where ${\bf W} = \{\{{\bf U}^{[l-1,l]}\}_{l=2}^L, \{{\bf V}^{[l]}\}_{l=2}^L\}$ is the model parameters in PINN-1D. By minimizing a cost function
$
{\cal L}({\bf y}) = {\cal L}\Big(f\big({\bf x}, {\bf W}, f_{\beta}(K, {\bf W}_{\beta})\big)\Big)$
with back propagation algorithm \cite{rumelhart1986learning}, ${\bf W}$  and ${\bf W}_{\beta}$ can be optimized.

Both the training phase and operation phase can benefit from the architecture of PINN-1D-Adp-$K$. A PINN with small size can  be first trained using the samples generated in the scenarios with small values of $K$.  The training complexity can be reduced because only a small size DNN needs to be trained. Thanks to the FC-DNN-$\beta_K$, the trained PINN-1D-Adp-$K$ can operate in realistic scenarios where $K$ (say the number of users) changes over time.

In the following, we take predictive resource allocation and interference coordination as two examples to illustrate how to apply the PINN-1D and PINN-2D. Since the PINNs are applicable to different manners of supervision on training, we consider unsupervised learning for predictive resource allocation policy and consider supervised learning for interference coordination.

\section{Case Study I: Predictive Resource Allocation} \label{sec: case study I PRA}
In this section, we demonstrate how the optimal predictive resource allocation (PRA) policy is learned by PINN-1D. Since generating labels from numerically obtained solutions is with prohibitive complexity for learning the PRA policy, we consider unsupervised learning.

\vspace{-2mm}\subsection{Problem Statement}
\subsubsection{System Model}
Consider a cellular network with $N_{\rm b}$ cells, where each BS is equipped with $N_{\sf tx}$ antennas and connected to a central processor (CP). The BSs may serve both real-time traffic and non-real-time (NRT) traffic. Since real-time service is with higher priority, NRT traffic is served with residual resources of the network after the quality of real-time service is guaranteed.

We optimize the PRA policy for mobile stations (MSs) requesting NRT service, say requesting for a file. Suppose that $K$ MSs in the network initiate requests at the beginning of a prediction window, and the $k$th MS (denoted as MS$_k$) requests a file with $B_k$ bits.

Time is discretized into frames each with duration $\Delta$, and each frame includes $T_s$ time slots each with duration of unit time. The durations are defined according to the channel variation, i.e., the coherence time of large scale fading (i.e., path-loss and shadowing) and small scale fading due to user mobility.
The prediction window contains $T_f$ frames.

Assume that an MS is only associated to the BS with the highest average channel gain (i.e.,  large scale channel gain) in each frame. To avoid multi-user interference, we consider time division multiple access as an illustration, i.e., each BS serves only one MS with all residual bandwidth and transmit power after serving real-time traffic in each time slot, and serves multiple MSs in the same cell in different time slots. Then, maximal ratio transmission is the optimal beamforming. Assume that the residual transmit power is proportional to the residual bandwidth \cite{YTCOM2016}, then the achievable rate of  MS$_k$ in the $t$th time slot of the $j$th frame can be expressed as $R^{j,t}_k =  W^{j,t}\log_2\Big(1+\frac{\alpha^j_k \|{\bm \gamma}^{j,t}_k\|^2}{\sigma_0^2} P_{\max}\Big)$, where $W^{j,t}$ and $P_{\max}$ are respectively the residual bandwidth and the maximal transmit power in the $t$th time slot of the $j$th frame, $\sigma_0^2$ is the noise power, ${\bm \gamma}_{k}^{j,t} \in \mathbb{C}^{N_{\sf tx} \times 1}$ is the small scale channel vector with ${\mathbb E}\{{\bm \gamma}_{k}^{j,t}\}=N_{\sf tx}$, $\alpha^j_k$ is the large scale channel gain. When $N_{\sf tx}$ and $T_s$ are large, it is easy to show that the time-average rate in the $j$th frame of MS$_k$ can be accurately approximated as,
\begin{eqnarray}\label{R}	
    R^j_k  \triangleq  \frac{1}{T_s}\sum_{t=1}^{T_s} R^{j,t}_k = \frac{1}{T_s}\sum_{t=1}^{T_s} W^{j,t}\log_2\Big(1+\frac{\alpha^j_k \|{\bm \gamma}^{j,t}_k\|^2}{\sigma_0^2}P_{\max}\Big)\approx W^j\log_2\Big(1+\frac{\alpha^j_k N_{\sf tx}}{\sigma_0^2}P_{\max}\Big),
\end{eqnarray}
where $W^j=\frac{1}{T_s}\sum_{t=1}^{T_s}W^{j,t}$ is the time-average residual bandwidth in the $j$th frame.
The time-average rates of each MS in the frames of the prediction window can either be predicted directly \cite{NB2018} or indirectly by first predicting the trajectory of each MS \cite{LSTMtrjactory17} and the real-time traffic load of each BS \cite{wang2017spatiotemporal} and then translating to average channel gains and residual bandwidth \cite{Guo2018Exploiting}.
\subsubsection{Optimizing Predictive Resource Allocation Plan}
We aim to optimize a resource allocation plan that minimizes the total transmission time required to ensure the quality of service (QoS) of each MS. The plan for MS$_k$ is denoted as ${\bf s}_k=[s^1_k,\cdots,s^{T_f}_k]^{\sf T}$, where $s^j_k$ is the fraction of time slots assigned to the MS in the $j$th frame.

The objective function can be expressed as $\sum_{k=1}^K\sum_{j=1}^{T_f} s^j_k$. To guarantee the QoS, the requested file should be completely downloaded to the MS before an expected deadline. For  simplicity, we let the duration between the time instant when an MS initiates a request and the transmission deadline equals the duration of the prediction window. Then, the QoS constraint can be expressed as $\sum_{j=1}^{T_f}  s_k^{j} R_k^j/B_k\Delta=1$.

Denote $r^j_k \triangleq R_k^j/B_k\Delta$ and ${\bf r}_k=[r^1_k,\cdots,r^{T_f}_k]^{\sf T}$, which is called \emph{average rate} in the sequel.
Then, the optimization problem can be formulated as,
\begin{subequations}\label{P1-vector}
	\vspace{-1mm}
	\begin{align}
	{\bf P1}:\min_{\bf S} &~~~ \|{\bf S}\|_1 \\
	{\rm s.t.} &~~~{\bf S}^{\sf T}\cdot{\bf R}\star{\bf I}={\bf I},\label{P1-c-vec}\\
	&~~~{\bf S}\cdot{\bf M}_i^{\sf T}\star{\bf I} \preceq {\bf I}, i=1,\cdots,N_{\rm b},\label{P1-d-vec}\\
	&~~~{\bf S}\succeq {\bf 0},
	\end{align}
\end{subequations}
where ${\bf S} = [{\bf s}_1,\cdots, {\bf s}_K], {\bf R} = [{\bf r}_1,\cdots, {\bf r}_K]$,  $({\bf M}_i)_{jk}=1$ or 0 if MS$_k$ associates or not associates to the $i$th BS in the $j$th frame, $(\cdot)_{jk}$ stands for the element in the $j$th row and $k$th column of a matrix.  \eqref{P1-c-vec} is the QoS constraint, and \eqref{P1-d-vec} is the resource constraint that ensures the total time allocated in each frame of each BS  not exceeding one frame duration. In  \eqref{P1-c-vec} and \eqref{P1-d-vec}, ``$\cdot$'' denotes matrix multiplication, and ``$\star$'' denotes element wise multiplication, ${\bf A}\preceq{\bf B}$ and ${\bf A}\succeq{\bf B}$ mean that each element in ${\bf A}$ is not larger or smaller than each element in ${\bf B}$, respectively.

After the plan for each MS is made by solving \textbf{P1} at the start of the prediction window, a transmission progress can be computed according to the plan as well as the predicted average rates, which  determines how much data should be transmitted to each MS in each frame. Then, each BS schedules the MSs in its cell in each time slot, see details in \cite{YTCOM2016}.

\subsection{Unsupervised Learning for Resource Allocation Plan}
\textbf{P1} is a convex optimization problem, which can be solved by interior-point method. However, the computational complexity scales with ${\cal O}(KT_f)^{3.5}$, which is prohibitive. To reduce on-line computational complexity, we can train a DNN to learn the optimal resource allocation plan. To avoid the computational complexity in generating labels, we train the DNN with unsupervised learning. To this end, we transform \textbf{P1} into a functional optimization problem as suggested in \cite{sun2019pimrc}. In particular, the relation between the optimal solution of \textbf{P1} and the known parameters (denoted as ${\bf S}({\bm \theta})$) can be found from the following problem as proved in \cite{sun2019pimrc},
\vspace{-1mm}
\begin{subequations}\label{P2-vector}
	\begin{align}
	{\bf P2}:\min_{\bf S({\bm \theta})} &~~~ {\mathbb E}_{\bm \theta}\{\|{\bf S}({\bm \theta})\|_1\} \\
	{\rm s.t.} &~~~{\bf S({\bm \theta})}^{\sf T}\cdot{\bf R}\star{\bf I}={\bf I},\label{P2-c-vec}\\
	&~~~{\bf S({\bm \theta})}\cdot{\bf M}_i^{\sf T}\star{\bf I} \preceq {\bf I}, i=1,\cdots,N_{\rm b},\label{P2-d-vec}\\
	&~~~{\bf S({\bm \theta})}\succeq {\bf 0} \label{P2-e-vec},
	\end{align}
\end{subequations}
where ${\bm \theta} = \{{\bf R}, {\bf M}_1, \cdots, {\bf M}_{N_{\rm b}}\}$ are the known parameters.

Problem \textbf{P2} is convex, hence it is equivalent to its Lagrangian dual problem \cite{convexopt},
\vspace{-1mm}
\begin{subequations}\label{P3}
	\begin{align}
	{\bf P3}:\notag\\\max_{\bm \lambda(\bm \theta)}\min_{\bf S({\bm \theta})} &{\cal L}\triangleq
	{\mathbb E}_{\bm \theta}\Bigg\{\|{\bf S}({\bm \theta})\|_1\!
	+\!{\bm\mu}^{\sf T}({\bm \theta})\big({\bf S({\bm \theta})}^{\sf T}\cdot{\bf R}\star{\bf I}-{\bf I}\big)\!\cdot\!{\bm 1}
	+\sum_{i=1}^{N_{\rm b}}{\bm\nu}_i^{\sf T}({\bm \theta})\big({\bf S({\bm \theta})}\cdot{\bf M}_i^{\sf T}\star{\bf I}- {\bf I}\big)\!\cdot\!{\bm 1}\notag\\
	&\hspace{10.5cm}-{\bm\Upsilon}({\bm \theta})\star{\bf S}({\bm \theta})\Bigg\}\label{P3-o} \\
	{\rm s.t.} &~~~{\bm\Upsilon}({\bm \theta})\succeq {\bm 0},{\bm\nu}_i(\bm \theta)\succeq {\bm 0}, \forall i\in\{1,\cdots,N_{\rm b}\},\label{P3-c-vec}
	\end{align}
\end{subequations}
where ${\cal L}$ is the Lagrangian function, ${\bm\lambda}(\bm \theta)=\{{\bm\mu}(\bm \theta), {\bm\nu}_1(\bm \theta),\cdots, {\bm\nu}_{N_{\rm b}}(\bm \theta),{\bm\Upsilon}({\bm \theta})\}$ is the set of Lagrangian multipliers. Considering the universal approximation theorem \cite{Hornik1989UnivApprox}, ${\bf S}({\bm \theta})$ and ${\bm\lambda}(\bm \theta)$ can be approximated with DNN  \cite{sun2019pimrc}.

\subsubsection{Design of the DNN} \label{sec: DNN design}
The input of a DNN to learn ${\bf S}({\bm \theta})$ can be designed straightforwardly as ${\bm \theta} = \{{\bf R}, {\bf M}_1, \cdots, {\bf M}_{N_{\rm b}}\}$, which is of high dimension. To reduce the input size, consider the fact that to satisfy constraint \eqref{P2-d-vec}, we can learn the resource allocated by each BS with a neural network (called DNN-$s$), because the resource conflictions only exist among the MSs associated to the same BS. In this way, the input only contains the known parameters of a single BS instead of all the BSs in the network.

The input of DNN-$s$ is ${\bf x}_i={\rm vec}({\bf R} \star {\bf M}_i)=[({\bf x}_{1,i})^{\sf T}, \cdots, ({\bf x}_{K,i})^{\sf T}]^{\sf T}$, where ${\rm vec}(\cdot)$ denotes the operation of concatenating the columns of a matrix into a vector, ${\bf x}_{k,i} = [x_{k,i}^1, \cdots, x_{k, i}^{T_f}]^{\sf T}$, $x_{k,i}^j=r_k^j$ is the average rate of MS$_k$ if it is served by the $i$th BS in the $j$th frame, and $x_{k,i}^j=0$ otherwise.
The output of DNN-$s$ is the resource allocation plan of all the MSs when they are served by the $i$th BS, which is normalized by the total resources allocated to each MS to meet the constraint in \eqref{P2-c-vec}, i.e.,
$
\textstyle\hat{s}^j_k\!=\!\frac{\hat{s}^{j'}_{k} r^j_k}{\sum_{\tau=1}^{T_f}\hat{s}^{\tau'}_{k}r^{\tau}_k} \Big/r^j_k\!=\! \frac{\hat{s}^{j'}_{k}}{\sum_{\tau=1}^{T_f}\hat{s}^{\tau'}_{k}r^{\tau}_k},k\!=\!1,\!\cdots\!,K, j\!=\!1,\!\cdots\!,T_f$,
where $\hat{s}^{j'}_{k}$ and $\hat{s}^j_k$ are respectively the output of DNN-$s$ before and after normalization.
We use the commonly used \texttt{Softplus} (i.e., $y=g(x)\triangleq\log(1+\exp(x))$) as the activation function of the hidden layers and output layer to ensure the learned plan being equal or larger than $0$.

Since DNN-$s$ is used to learn ${\bf S}({\bm \theta})$ that is permutation invariant to ${\bf x}_i$, we can apply PINN-1D-Adp-$K$ whose input-output relation is $f_s({\bf x}_i, {\bf W}_s)$, where ${\bf W}_s$ denotes the model parameters in DNN$-s$. Both the input and output sizes of  DNN-$s$ are $KT_f$, which may change since the number of MSs may vary over time.

To learn the Lagrange multipliers, we design a FC-DNN called DNN-$\lambda$,  whose input-output relation is $f_{\nu}(\tilde{\bf x}_i, {\bf W}_{\nu})$. Since the constraint in \eqref{P2-c-vec} is already satisfied due to the normalization operation in the output of DNN-$s$ and the constraint \eqref{P2-e-vec} is already satisfied  due to the \texttt{Softplus} operation in the output layer of DNN-$s$, we do not need to learn multiplier ${\bm \mu}$ and ${\bm \Upsilon}$ in \eqref{P3-o} and hence we only learn multiplier ${\bm \nu}_i$. Since ${\bm \nu}_i$ is used to satisfy constraint \eqref{P2-d-vec}, which depends on ${\bf x}_i$, the input of DNN-$\lambda$ contains ${\bf x}_i$. Since the vector ${\bf x}_i$ is composed of the average rates of $K$ MSs, its dimension may vary with $K$. Since DNN-$\lambda$ is a FC-DNN whose architecture cannot change with $K$, we consider the maximal number of MSs $K_{\max}$ such that $K\leq K_{\max}$. Then, the input of DNN-$\lambda$ is $\tilde{\bf x}_i=[({\bf x}_{1,i})^{\sf T}, \cdots, ({\bf x}_{K_{\max},i})^{\sf T}]^{\sf T}$. When $K<K_{\max}$, ${\bf x}_{K_{\max},i}={\bf 0}$ for $\forall k>K$. The activation functions in hidden layers and output layer are \texttt{Softplus} to ensure the Lagrange multipliers being equal or larger than $0$, hence \eqref{P3-c-vec} can be satisfied.

\subsubsection{Training Phase}

DNN-$s$ and DNN-$\lambda$ are trained in multiple epochs, where in each epoch ${\bf W}_s$ and ${\bf W}_{\nu}$ are consecutively updated using the gradients of a cost function with respective to ${\bf W}_s$ and ${\bf W}_{\nu}$ via back-propagation. The cost function is the empirical form of \eqref{P3-o}, where ${\bf S}({\bm \theta})$ and ${\bm \lambda}({\bm \theta})$ are replaced by $f_s({\bf x}_i, {\bf W}_s)$ and $f_{\nu}({\bf x}_i, {\bf W}_{\nu})$. In particular, we replace ${\mathbb E}_{\bm\theta}\{\cdot\}$ in the cost function with empirical mean, because the probability density function of ${\bm\theta}$ is unknown. We omit the second and third term in \eqref{P3-o} because the constraint \eqref{P2-c-vec} and \eqref{P2-e-vec} can be ensured by the normalization and \texttt{Softplus} operation in the output of DNN-$s$, respectively. Moreover, we add the cost function with an augmented Lagrangian term \cite{hestenes1969multiplier} to make the learned policy to satisfy the constraints in \textbf{P2}. The cost function is expressed as,
\vspace{-2mm}
\begin{eqnarray}
\hat{\cal L}({\bf W}_s, {\bf W}_{\nu}) = \frac{1}{N} \sum_{n=1}^N \sum_{i=1}^{N_{\rm b}}\Bigg( \left\|{\bm f}_{s,i}^{(n)}\right\|_1 &+& ({\bm f}_{\nu,i}^{(n)})^{\sf T}\big( [{\bm f}_{s,i}^{(n)}]_{T_f\times {K_{\max}}} \cdot {\bf M}_i^{\sf T} \star {\bf I} - {\bf I}\big)\cdot{\bm 1}\notag\\
&+& \underbrace{\frac{\rho}{2}\left\|\big( [{\bm f}_{s,i}^{(n)}]_{T_f\times {K_{\max}}} \cdot {\bf M}_i^{\sf T} \star {\bf I} - {\bf I}\big)^+\cdot{\bm 1}\right\|^2}_{(a)}\Bigg),\notag \nonumber
\end{eqnarray}
where $N$ is the number of training samples, ${\bm f}_{s,i}^{(n)}\triangleq f_s({\bf x}_i^{(n)}, {\bf W}_s)$, ${\bm f}_{\nu,i}^{(n)}\triangleq f_{\nu}(\tilde{\bf x}_i^{(n)}, {\bf W}_{\nu})$, ${\bf x}^{(n)}_i$ and $\tilde{\bf x}^{(n)}_i$ denote the $n$th sample of DNN-$s$ and DNN-$\lambda$, respectively, $[{\bf a}]_{m\times n}$ is the operation to represent vector ${\bf a}$ as a matrix with $m$ rows and $n$ columns, $(a)$ is the augmented Lagrangian term, which is a quadratic punishment for not satisfying the constraints. $(x)^+=x$ when $x\geq 0$ and $(x)^+=0$ otherwise, $\rho$ is a parameter to control the punishment. It is proved in \cite{hestenes1969multiplier} that the optimality can be achieved as long as $\rho$ is larger than a given value. Hence we can regard $\rho$ as a hyper-parameter.

In DNN-$s$, ${\bf W}_s$ is trained to minimize $\hat{\cal L}({\bf W}_s, {\bf W}_{\nu})$. In DNN-$\lambda$, ${\bf W}_{\nu}$ is trained to maximize $\hat{\cal L}({\bf W}_s, {\bf W}_{\nu})$. The learning rate is adaptively updated with Adam algorithm \cite{Kingma2014Adam}.
\subsubsection{Operation Phase}
For illustration, assume that $\bf R$ and ${\bf M}_i, i=1,\cdots,N_{\rm b}$ are known at the beginning of the prediction window. Then, by sequentially inputting the trained DNN-$s$ with ${\bf x}_i={\rm vec}({\bf R} \star {\bf M}_i), i=1,\cdots,N_{\rm b}$, DNN-$s$ can sequentially output the resource allocation plans for all MSs served by the $1,\cdots,N_{\rm b}$th BS.

\section{Case Study II: Interference Coordination} \label{sec: case study II IC}
In this section, we demonstrate how an interference coordination policy considered in \cite{sun2017learning} is learned by PINN-2D. For a fair comparison, we consider supervised learning as in \cite{sun2017learning}.

Consider a wireless interference network with $K$ single-antenna transmitters and $K$ single-antenna receivers, as shown in Fig. \ref{fig:fig-wlchnl}. To coordinate interference among links, the power at each transmitter is controlled to maximize the sum-rate as follows,\vspace{-2mm}
\begin{subequations}\label{P: max sum-rate}
	\begin{align}
	\max_{p_1,\cdots, p_K} ~~& \sum_{k=1}^K \log\left(1+\frac{|\gamma_{kk}|^2p_k}{\sum_{n=1,n\neq k}^K|\gamma_{nk}|^2 + \sigma_0^2}\right) \label{P: msr-1} \\
	{\rm s.t.} ~~& 0\leq p_k \leq P_{\max}, \forall k=1,\cdots,K, \label{P: msr-2}
	\end{align}
\end{subequations}\vspace{-1mm}
where $\gamma_{mn}\in {\mathbb C}$ is the channel between the $m$th transmitter and the $n$th receiver, $m,n=1,\cdots,K$, $P_{\max}$ is the maximal transmit power of each transmitter, and $\sigma_0^2$ is the noise power.

\vspace{-1mm}Problem \eqref{P: max sum-rate} is NP-hard, which can solved numerically by a weighted-minimum-mean-squared-error (WMMSE) algorithm\cite{sun2017learning}.\vspace{-1mm}

We use PINN-2D-Adp-$K$ to learn the power control policy. The input of the DNN is the channel matrix, i.e.,
\vspace{-2mm}\begin{equation}\label{eq: Xinput}
{\bf X} = \left[
\begin{tabular}{ccc}
$|\gamma_{11}|$ & $\cdots$ & $|\gamma_{1K}|$ \\
$\vdots$ & $\ddots$ & $\vdots$ \\
$|\gamma_{K1}|$ & $\cdots$ & $|\gamma_{KK}|$,
\end{tabular}
\right],
\end{equation}
and the output is the transmit power normalized by the maximal transmit power, i.e., ${\bf y}=[p_1,\cdots,p_K]^{\sf T}/P_{\max}$. Then, the constraint \eqref{P: msr-2} becomes ${\bf 0}\preceq {\bf y}\preceq {\bf 1}$. The expected output of the DNN (i.e., the label) is the solution obtained by WMMSE algorithm that is also normalized by $P_{\max}$), i.e., ${\bf y}^*=[p_{1}^*,\cdots,p_{K}^*]^{\sf T}/P_{\max}$.
The activation function of the hidden layers is the commonly used {\tt Softplus} and the activation function of the output layer is {\tt Sigmoid} (i.e., $y=1/(1+e^{-x})$) such that ${\bf 0}\preceq {\bf y}\preceq {\bf 1}$, hence constraint \eqref{P: msr-2} can be guaranteed. We add batch normalization in the output layer to avoid gradient vanishing \cite{ioffe2015batch}.

The model parameters ${\bf W}=\{{\bf A}^{[l-1,l]}, {\bf B}^{[l-1,l]}, {\bf C}^{[l-1,l]}, {\bf D}^{[l-1,l]}\}_{l=2}^L$ are trained to minimize the empirical mean square errors between the outputs of the DNN and the expected outputs over $N$ training samples. Each sample is composed of a randomly generated channel matrix as in \eqref{eq: Xinput} and the corresponding solution obtained from the WMMSE algorithm.

\section{Simulation Results} \label{sec: simulation results}
In this section, we evaluate the performance of the proposed solutions. We consider the two tasks in previous case studies, which are respectively 1D- and 2D-permutation invariant.

All simulations are implemented on a computer with one 14-core Intel i9-9940X CPU, one Nvidia RTX 2080Ti GPU, and 64 GB memory. The optimal solution of PRA is implemented in Matlab R2018a with the build-in interior-point algorithm, and the WMMSE algorithm is implemented in Python 3.6.4 with the open-source code of \cite{sun2017learning} from Github (available: \url{https://github.com/Haoran-S/SPAWC2017}). The training of the DNNs is implemented in Python 3.6.4 with TensorFlow 1.14.0.

\vspace{-3mm}\begin{figure}[!htp]
	\centering
	\includegraphics[width=0.8\linewidth]{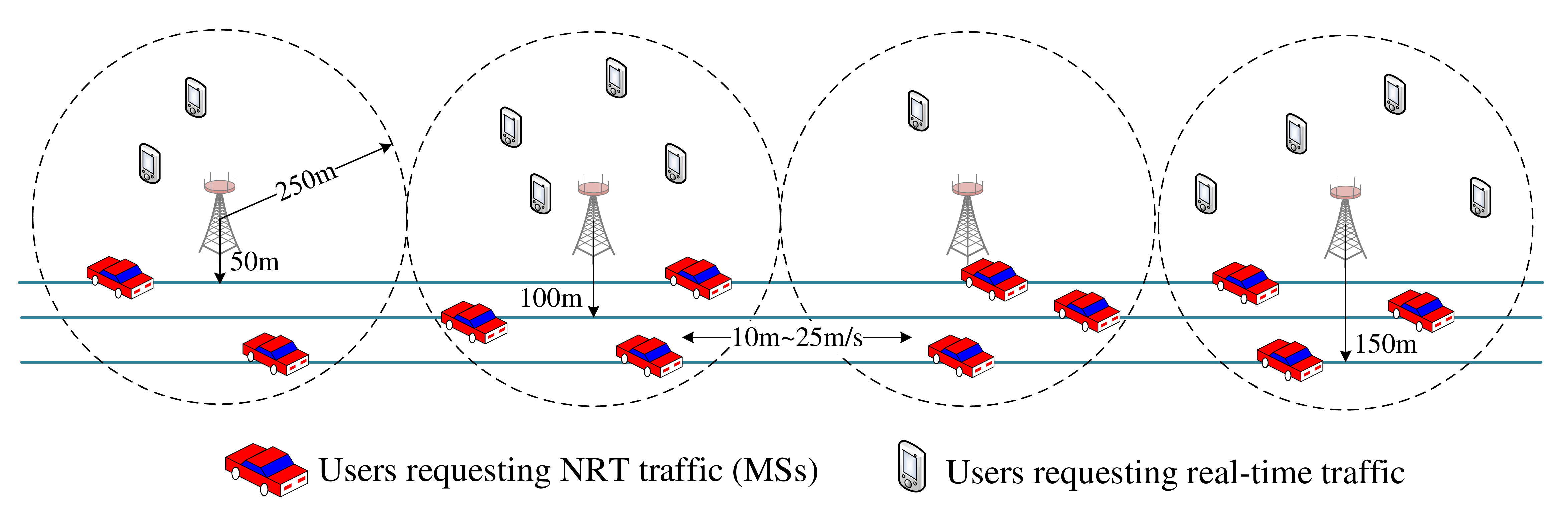}\vspace{-4mm}
	\caption{Simulation setup for predictive resource allocation to $K$ NRT users, $K$ randomly changes from 1 to $K_{\max}=40$.}
	\label{fig:setup}
\end{figure}\vspace{-2mm}

\vspace{-1mm}\subsection{Predictive Resource Allocation} \label{sec: simu-PRA}
\vspace{-1mm}\subsubsection{Simulation Setups}
Consider a cellular network with cell radius $R_{\rm b}=250$ m, where four BSs each equipped with $N_{\sf tx}=8$ antennas are located along a straight line. For each BS, $P_{\max}$ is 40 W, $W_{\max}=20$ MHz and the cell-edge SNR is set as 5 dB, where the intercell interference is implicitly reflected. The path loss model is $36.8 + 36.7\log_{10}(d)$, where $d$ is the distance between the BS and MS in meter. The MSs move along three roads of straight lines with minimum distance from the BSs as $50$ m, $100$ m and $150$ m, respectively. At the beginning of the prediction window, $K$ MSs at different locations in the roads initiate requests, where each MS requests a file with size of $B_k=6$ Mbytes (MB).
Each frame is with duration of $\Delta=1$ second, and each time slot is with duration $10$ ms, i.e., each frame contains $T_s=100$ time slots.

To characterize the different resource usage status of the BSs by serving the real-time traffic, we consider two types of BSs:  busy BS  with average residual bandwidth in the prediction window $\overline{W}= 5$ MHz and idle BS with $\overline{W}= 10$ MHz, which are alternately located along the line as idle, busy, idle, busy, as shown in Fig. \ref{fig:setup}. The results are obtained from 100 Monte Carlo trials. In each trial, $K$ is randomly selected from 1 to $K_{\max}=40$, the MSs initiate requests randomly at a location along the trajectory, and travel with speed uniformly distributed in $(10, 25)$ m/s and directions uniformly selected from 0 or +180 degree. The small-scale channel in each time slot changes independently according to Rayleigh fading, and the residual bandwidth at each BS in each time slot varies according to Gaussian distribution with mean value $\overline{W}$ and standard derivation $0.2\overline{W}$. The setup is used in the sequel unless otherwise specified.

Each sample for unsupervised training or for testing is generated as follows. For the $K$ MSs, the indicator of whether a MS is served by the $i$th BS, ${\bf M}_i$, can be obtained. The average channel gains of the MSs are computed with the path loss model. With the simulated residual bandwidth in each BS, the average rates of $K$ users within the prediction window, ${\bf R}$, can be computed with \eqref{R}. Then, a sample can be obtained as ${\bf x}_i={\rm vec}({\bf R} \star {\bf M}_i),i=1,\cdots,N_{\rm b}$.

As demonstrated previously, the architecture of PINN can be flexibly controlled to adapt to different values of $K$, with which the training complexity can be further reduced. In order to show the complexity reduction respectively brought by the network size adaptation and by the parameter sharing among blocks, we train three different kinds of DNN-$s$ as follows, each of them is trained together with a DNN-$\lambda$.
\begin{itemize}
	\item \emph{PINN-1D-Adp-$K$:} This DNN-$s$ is with the architecture in Fig. \ref{fig:dnn-dimgen}. The training samples are generated in the scenarios with different number of MSs, where the majority of the samples are generated when $K$ is randomly selected from $1 \sim 10$ and the rest of 1000 samples are generated when $K=K_{\max}=40$.
	
	\item \emph{PINN-1D:} This DNN-$s$ is with the architecture in Fig. \ref{fig:dnn}. The training samples are generated by a simulated system with $K=K_{\max}$ users.
	
	\item \emph{FC-DNN:} This DNN-$s$ is the FC-DNN without parameter sharing, which is with the same number of layers and the same number of neurons with PINN-1D. The training samples are also generated in the scenario where $K=K_{\max}$.
\end{itemize}

The fine-tuned hyper-parameters for these DNNs when $T_f=60$ seconds and $K_{\max}=40$ are summarized in Table \ref{table}. When $T_f$ changes, the hyper-parameters should be tuned again to achieve the best performance. The training set contains 10,000 samples and the test set contains 100 samples, where the testing samples are generated in the scenario where $K=K_{\max}=40$.

\begin{table}[htb!]
	\centering
	\vspace{-2mm}
	\caption{Hyper-parameters for the DNNs When $T_f=60$ Seconds and $K_{\max}=40$.}\label{table}
	\vspace{-2mm}
	\small
	\begin{tabular}{c|c|c|c|c|c}
		\hline\hline
		\multirow{2}{*}{\textbf{Parameters}} & \multicolumn{4}{c}{\textbf{Values}} \\
		\cline{2-6}
		~ & \multicolumn{2}{c|}{PINN-1D-Adp-$K$} &\multirow{2}{*}{PINN-1D}&\multirow{2}{*}{FC-DNN}& \multirow{2}{*}{DNN-$\lambda$}  \\
		\cline{2-3}
		~ & PINN-1D & FC-DNN-$\beta_K$ & ~ & ~ & ~\\
		\hline
		\tabincell{c}{Number of input nodes} & $KT_f$ & 1 &$K_{\max}T_f = 2400$&$K_{\max}T_f = 2400$& $K_{\max}T_f = 2400$ \\
		\hline
		\tabincell{c}{Number of hidden layers} & 2 & 1 &2&2& 2 \\
		\hline
		\tabincell{c}{Number of hidden nodes} &50$K$, 50$K$&10&2,000, 2,000&2000, 2000& 200, 100 \\
		\hline
		\tabincell{c}{Number of output nodes} &  $KT_f$ & 1 &$K_{\max}T_f=2400$&$K_{\max}T_f=2400$&  $T_f = 60$ \\
		\hline
		\tabincell{c}{Initial learning  rate} & \multicolumn{5}{c}{0.01} \\
		\hline
        \tabincell{c}{Learning algorithm} & \multicolumn{5}{c}{Adam} \\
 		\hline
 		\tabincell{c}{Back propagation algorithm} & \multicolumn{5}{c}{Iterative batch gradient descent \cite{goodfellow2016deep}} \\
		\hline\hline
	\end{tabular}
\vspace{-5mm}
\end{table}
\subsubsection{Number of Model Parameters}
In PINN-1D, the weight matrix ${\bf W}^{[l-1,l]}$ contains two sub-matrices ${\bf U}^{[l-1,l]}$ and ${\bf V}^{[l-1,l]}$, each of which contains $N^{[l-1,l]}$ model parameters. Hence, the total number of model parameters is $2\sum_{l=2}^LN^{[l-1,l]}$.

In PINN-1D-Adp-$K$, the number of model parameters is $2\sum_{l=2}^LN^{[l-1,l]}+\sum_{l=2}^LN_{\beta}^{[l-1,l]}$, where the first and second term respectively correspond to the model parameters in PINN-1D and FC-DNN-$\beta_K$, $N_{\beta}^{[l-1,l]}$ is the number of parameters in the weights between the $(l-1)$th and $l$th layer of FC-DNN-$\beta_K$.
For the PINN-1D with hyper-parameters in Table \ref{table}, the input contains $K_{\max}=40$ blocks and each block contains $T_f=60$ elements, the first hidden layer also contains $K_{\max}=40$ blocks and each block contains $2,000/K_{\max}=50$ elements. Then, $N^{[1,2]}=60\times 50=3,000$. Similarly, $N^{[2,3]}=50\times 50=2,500$, and $N^{[3,4]}=50\times 60=3,000$. Hence, there are $2\times(3,000+2,500+3,000)=17,000$ model parameters in PINN-1D. For FC-DNN-$\beta_K$  with hyper-parameters in Table \ref{table}, $N_{\beta}^{[1,2]}=1\times 10=10$ and $N_{\beta}^{[2,3]}=10\times 1=10$, hence there are $\sum_{l=2}^LN_{\beta}^{[l-1,l]}=20$ model parameters, which is with much smaller size than PINN-1D.

In the FC-DNN with the same number of hidden layers and the same number of neurons in each hidden layer as PINN-1D, the number of parameters in ${\bf W}^{[l-1,l]}$ is $K_{\max}^2\sum_{l=2}^LN^{[l-1,l]}$, which is $K_{\max}^2/2$ as large as PINN-1D. For the FC-DNN  with hyper-parameters in Table \ref{table}, the number of model parameters is $40^2\times(3,000+2,500+3,000)=13,600,000$, which increases by $K_{\max}^2/2=800$ times over PINN-1D.

\subsubsection{Sample and Computational Complexity} \label{sec: PRA:complexities}
\emph{Sample complexity} is defined as the minimal number of training samples for a DNN to achieve an expected performance, and \emph{computational complexity} is measured by the running time consumed by training the DNNs.

\begin{figure}[!htb]
	\centering
	\begin{minipage}[t]{0.45\linewidth}	
		\subfigure[Sample complexity]{
			\includegraphics[width=\textwidth]{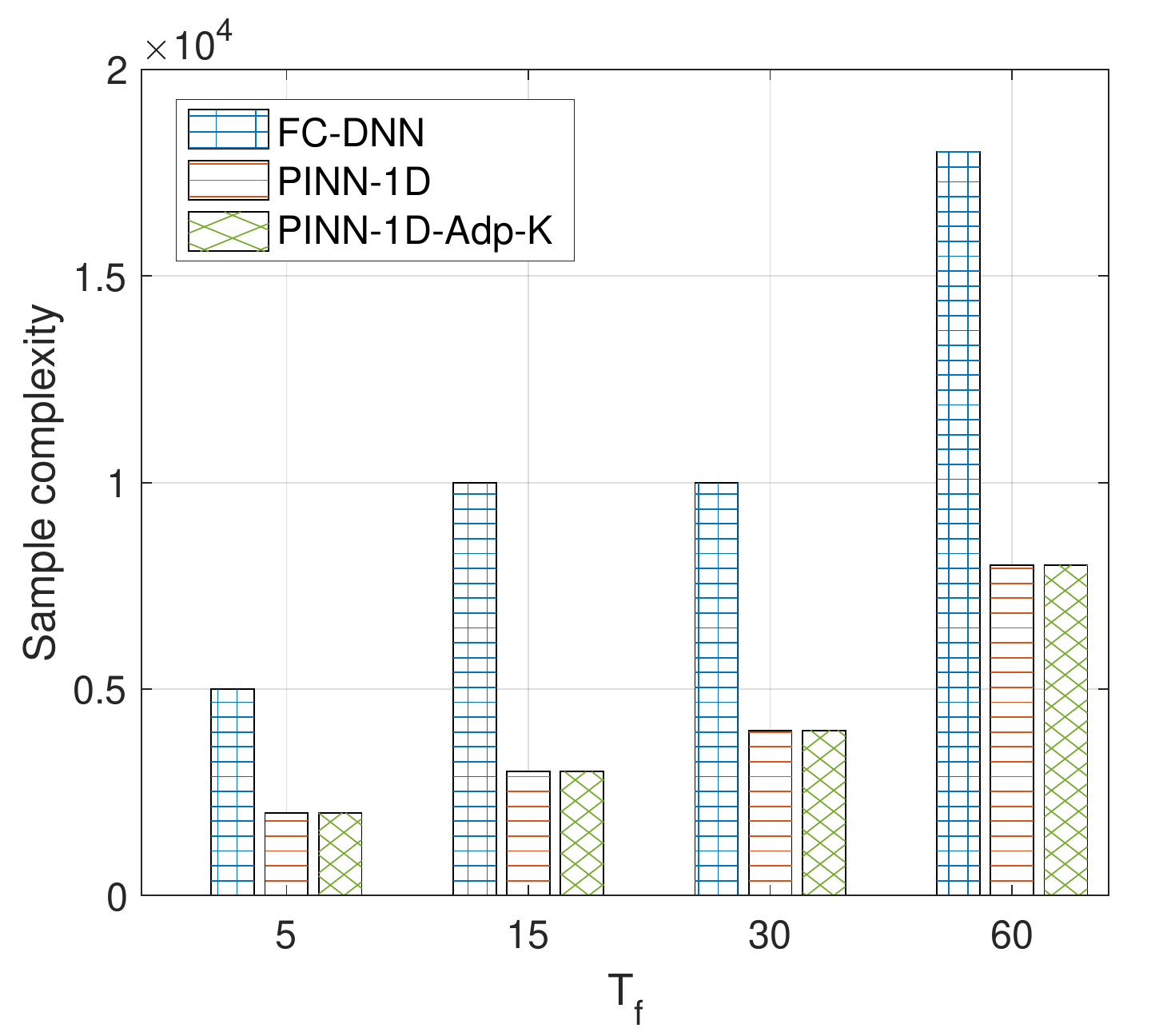}}
	\end{minipage}
	\begin{minipage}[t]{0.45\linewidth}	
		\subfigure[Computational complexity]{
			\includegraphics[width=\textwidth]{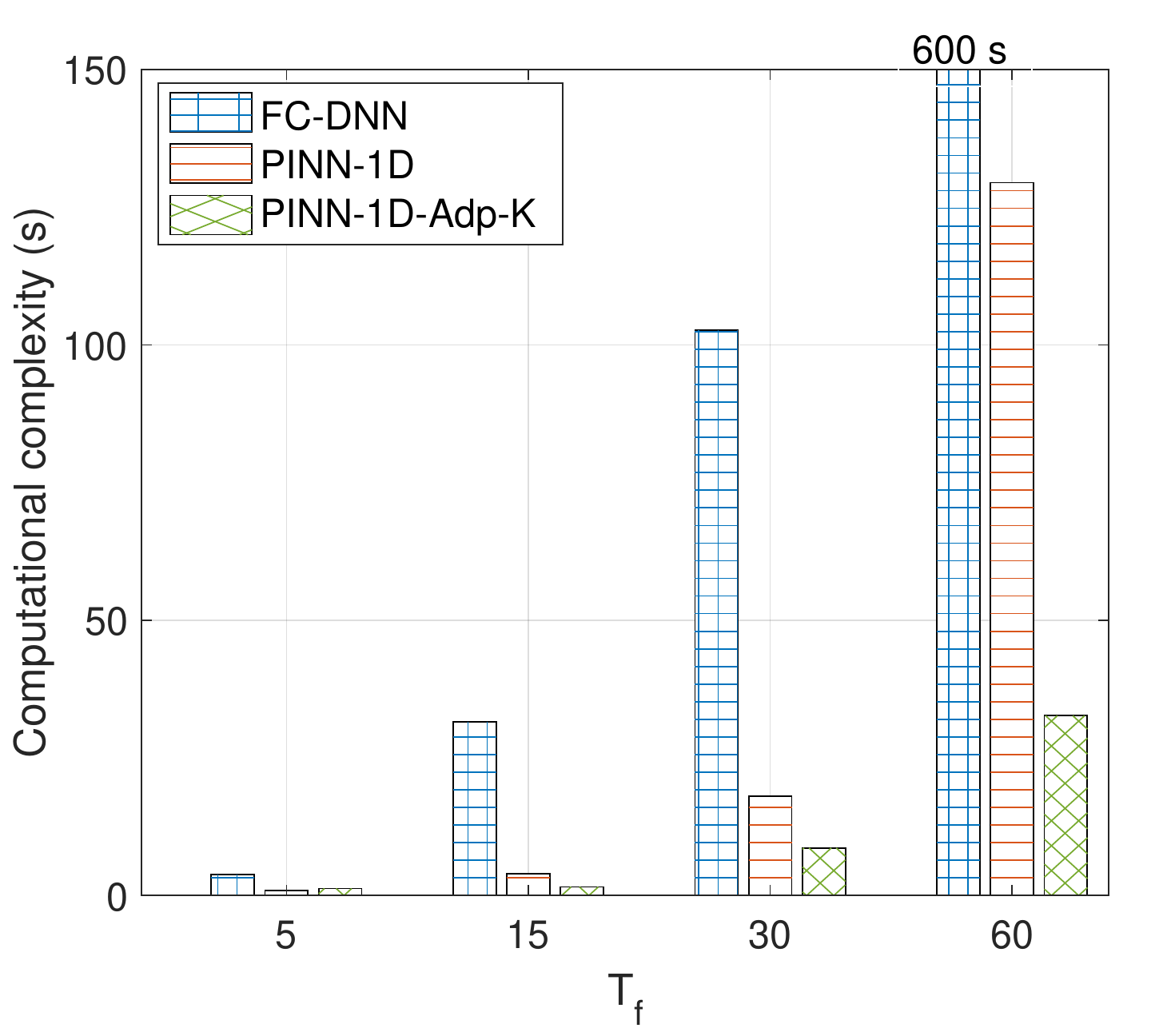}}
	\end{minipage}\vspace{-2mm}
	\caption{Training complexities of DNNs for learning PRA.}\label{fig: PRA-cplxty}\vspace{-5mm}
\end{figure}

In Fig. \ref{fig: PRA-cplxty}, we provide the sample and computational complexities of all the DNNs when the objective in \textbf{P1} on the test set can achieve less than 20\% performance loss from the optimal value (i.e., the total allocated time resource for all MSs), which is obtained by solving \textbf{P1} with interior-point method. In Fig. \ref{fig: PRA-cplxty} (b), when $T_f=60$ s, the computational complexity of training FC-DNN is 600 s, which is out of the range of $y$-axis.

\vspace{-1mm}

We can see that the training complexities  of PINN-1D and PINN-1D-Adp-$K$ are much lower than ``FC-DNN'', because the PINNs can converge faster thanks to the reduced model parameters by parameter sharing. The computational complexity of PINN-1D-Adp-$K$ is lower than PINN-1D due to the less number of neurons in each layer during the training phase. The computational complexity reduction of  PINN-1D and PINN-1D-Adp-$K$ from ``FC-DNN'' grows with $T_f$. When $T_f=5$ s, the computational complexity of PINN-1D-Adp-$K$ is 67\% less than ``FC-DNN'', while when $T_f=60$ s, the complexity is reduced by 94\%. The sample complexities of the two PINNs are comparable, since their numbers of model parameters are comparable.

It is noteworthy that although DNN-$\lambda$ is not with parameter sharing, the training complexity of PINNs is still much lower by only applying parameter sharing to DNN-$s$. This is because the fine-tuned DNN-$\lambda$ has much less hidden and output nodes, as shown in Table \ref{table}.

\subsubsection{Performance of PRA Learned with DNNs}
To evaluate the dimensional generalization ability of PINN-1D-Adp-$K$, we compare the total transmission time required for downloading the files averaged over all MSs with the following methods.
\begin{itemize}
	\item \textbf{Proposed-1}: The resource allocation plan is obtained by the well-trained PINN-1D-Adp-$K$ with unsupervised learning. The training set contains 16000 samples, which are generated in scenarios where the numbers of MSs $K$ change randomly from 1 to 10.
	\item \textbf{Proposed-2}: The only difference from ``Proposed-1'' lies in the training set, where we add 2000 training samples generated from the scenario with $K=K_{\max}=40$ in addition to 14000 samples generated with $K \in [1,10]$.
	\item \textbf{Supervised}: The resource allocation plan is obtained by the PINN-1D trained in the supervised manner, where the labels in the training samples are generated by solving \textbf{P1} with interior-point method.
	\item \textbf{Optimal}: The resource allocation plan is obtained by solving \textbf{P1} with interior-point method.
	\item \textbf{Baseline}: This is a non-predictive method \cite{su2015user}, where each BS serves the MS with the earliest deadline in each time slot. If several MSs have the same deadline, then the MS with most bits to be transmitted is served firstly.
\end{itemize}

\begin{figure}[!htb]\vspace{-2mm}
	\centering
	\vspace{-2mm}
	\includegraphics[width=0.45\linewidth]{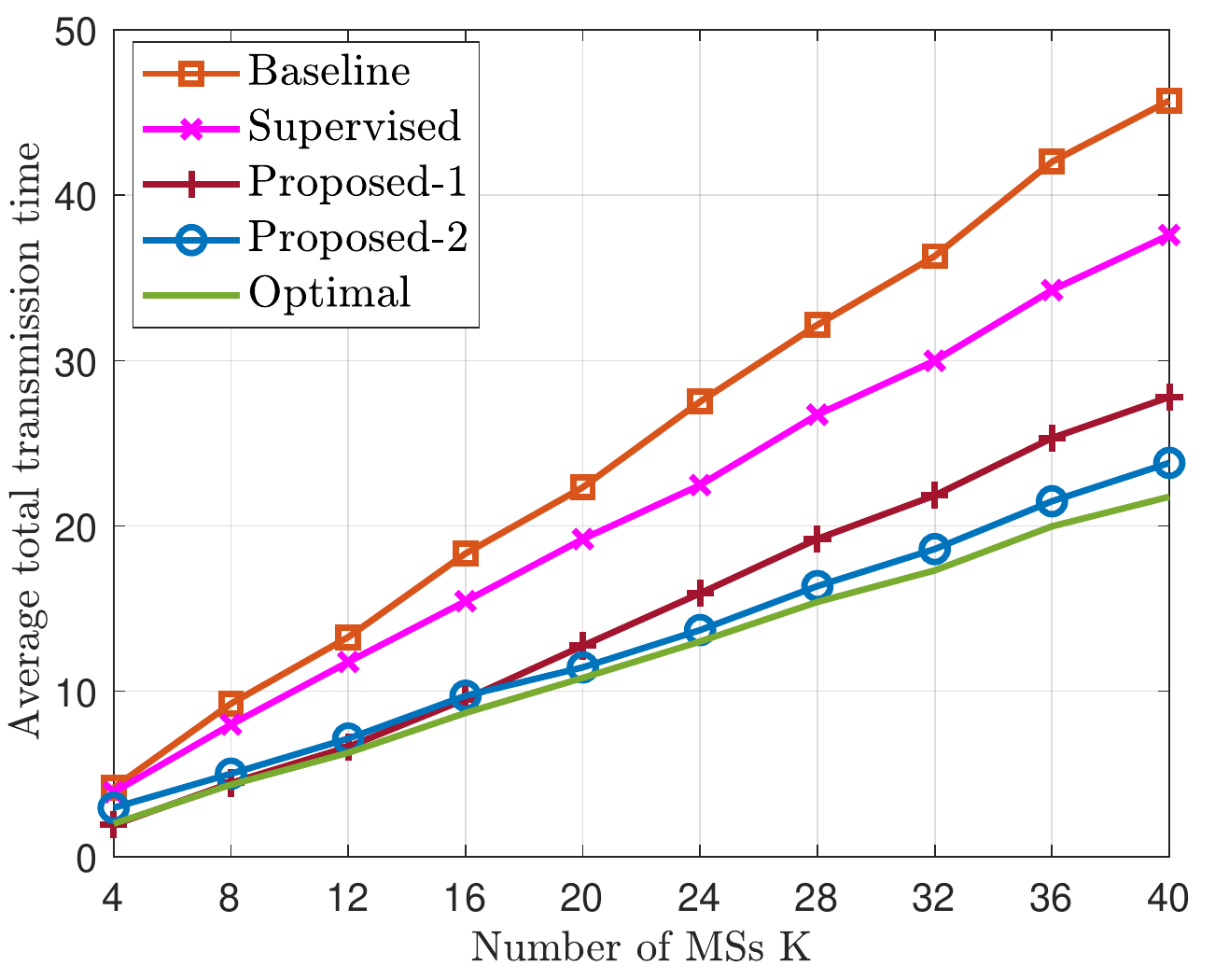}
	\vspace{-3mm}
	\caption{Performance comparison of all methods, $T_f=60$ seconds, $K_{\max}=40$.}
	\vspace{-5mm}
	\label{fig:figtimearr}\vspace{-2mm}
\end{figure}

In Fig. \ref{fig:figtimearr}, we provide the average total transmission time required for downloading a file. We can see that ``Proposed-1'' performs closely to the optimal method when $K$ is less than 20, but the performance loss is larger when $K$ is large. Nonetheless, by adding some training samples generated with large value of $K$ to learn $\beta_K=f_{\beta}(K, {\bf W}_{\beta})$, ``Proposed-2'' performs closely to the optimal method, while the training complexity keeps small as shown in previous results. Besides, the proposed methods with unsupervised DNN outperforms the method with supervised DNN. This is because the resource allocation plan learned from labels cannot satisfy the constraints in problem \textbf{P1}, which leads to resource confliction among users. Moreover, all the PRA methods outperform the non-predictive baseline dramatically.
\vspace{-2mm}\vspace{-2mm}

\vspace{-2mm}\subsection{Interference Coordination} \label{sec: simu-IC}
\subsubsection{Simulation Setups} \label{sec:simu:IC:setup}
Consider a wireless network with $K$ transmitters and $K$ receivers each equipped with a single antenna, where $K \leq K_{\max}$.
A power control policy is obtained either by the WMMSE algorithm or by a trained DNN, as discussed in section \ref{sec: case study II IC}.

When training the DNN with supervision, the samples $\{\bf X, {\bf y}^*\}$ are generated via Monte Carlo trials. In each trial, the channel matrix $\bf X$ in \eqref{eq: Xinput} is firstly generated with Rayleigh distribution, and then the label is obtained as ${\bf y}^*=[p_{1}^*,\cdots,p_{K}^*]^{\sf T}/P_{\max}$ by solving problem \eqref{P: max sum-rate} with WMMSE algorithm. The test set contains 1,000 samples.

We compare the sample and training complexities of three different DNNs, i.e., PINN-2D-Adp-$K$, PINN-2D and FC-DNN. When training ``PINN-2D-Adp-$K$'', 80\% training samples are generated in the scenario when $K$ is small\footnote{When $K_{\max}=10$ or $20$, the majority of samples are generated in the scenarios with $K\leq 5$, and when $K_{\max}=30$, the majority of samples are generated with $K\leq 10$.} and 20\% samples are generated in the scenario when $K=K_{\max}$. The hyper-parameters of the three DNNs are as follows. When $K_{\max}\neq30$, the hyper-parameters need to be fine-tuned again to achieve the best performance.

\begin{table}[htb!]
	\centering
	\vspace{-2mm}
	\caption{Hyper-parameters for the DNNs When $K_{\max}=30$.}\label{table-wmmse}
	\vspace{-2mm}
	\small
	\begin{tabular}{c|c|c|c|c}
		\hline\hline
		\multirow{2}{*}{\textbf{Parameters}} & \multicolumn{4}{c}{\textbf{Values}} \\
		\cline{2-5}
		~ & \multicolumn{2}{c|}{PINN-2D-Adp-$K$} &\multirow{2}{*}{PINN-2D}&\multirow{2}{*}{FC-DNN}\\
		\cline{2-3}
		~ & PINN-2D & FC-DNN-$\beta_K$ & ~ & ~ \\
		\hline
		\tabincell{c}{Number of input nodes} & $K^2$ & 1 &$K_{\max}^2 = 900$&$K_{\max}^2 = 900$\\
		\hline
		\tabincell{c}{Number of hidden layers} & 2 & 1 & 2 & 3\\
		\hline
		\tabincell{c}{Number of hidden nodes} & $9K^2$ & 10 & $90\times 90$ & $400, 300, 200$ \\
		\hline
		\tabincell{c}{Number of output nodes} &  $K$ & 1 &$K_{\max}=30$&$K_{\max}=30$\\
		\hline
		\tabincell{c}{Initial learning  rate} & \multicolumn{3}{c|}{0.01} & \multicolumn{1}{c}{0.001} \\
		\hline
		\tabincell{c}{Learning algorithm} & \multicolumn{4}{c}{RMSprop \cite{HintonGenerative}}  \\
		\hline
		\tabincell{c}{Back propagation algorithm} & \multicolumn{4}{c}{Iterative batch gradient descent \cite{goodfellow2016deep}} \\
		\hline\hline
	\end{tabular}
	\vspace{-5mm}
\end{table}
\subsubsection{Number of Model Parameters} \label{sec: 1D-No.Params}
Since there are two weight matrices between the $(l-1)$th and the $l$th layer, each weight matrix contains two sub-matrices, and each sub-matrix contains $N^{[l-1,l]}$ weights, the number of model parameters in PINN-2D is $4\sum_{l=2}^L N^{[l-1,l]}$.

The number of parameters in PINN-2D-Adp-$K$ is $4\sum_{l=2}^LN^{[l-1,l]}+\sum_{l=2}^LN_{\beta}^{[l-1,l]}$, where the first and second term respectively correspond to the parameters in PINN-2D and FC-DNN-$\beta_K$.

For PINN-2D with hyper-parameters in Table \ref{table-wmmse}, the input contains $K_{\max}^2=900$ blocks and each block ${\bf x}_{mn}$ is a scalar, the second layer also contains $900$ blocks and each block ${\bf h}^{[2]}_{mn}$ is a $(90/K_{\max})\times (90/K_{\max})=3\times 3$ matrix. Recall the number of rows and columns of the sub-matrices defined in \eqref{weight mat1}, there are $N^{[1,2]}=1\times 3=3$ model parameters in each sub-matrix. Similarly, $N^{[2,3]}=3\times 3=9$ and $N^{[3,4]}=3\times 1=3$. Hence, there are in total $4\sum_{l=2}^L N^{[l-1,l]}=60$ model parameters in PINN-2D. The number of model parameters in FC-DNN-$\beta_K$ is 20,  hence there are $60+20=80$ model parameters in PINN-2D-Adp-$K$.

The FC-DNN with hyper-parameters in Table \ref{table-wmmse} contains $900\times 400 +400\times 300 + 300\times 200 + 200\times 30=546,000$ model parameters. Hence, PINN-2D and PINN-2D-Adp-$K$ can reduce the model parameters by $546,000/60=9,100$ and $546,000/80=6,825$ times with respect to FC-DNN, respectively.

\subsubsection{Sample and Computational Complexity}
In the following, we compare the training complexities for the PINNs to achieve an expected performance on the test set, which is set as the best performance that all the DNNs can achieve. When $K_{\max}=10, 20, 30$, the performance is 90\%, 85\%, 80\% of the sum-rate that the WMMSE algorithm can achieve, respectively.
\vspace{-1mm}\begin{figure}[!htb]
	\centering
	\begin{minipage}[t]{0.45\linewidth}	
		\subfigure[Sample complexity]{
			\includegraphics[width=\textwidth]{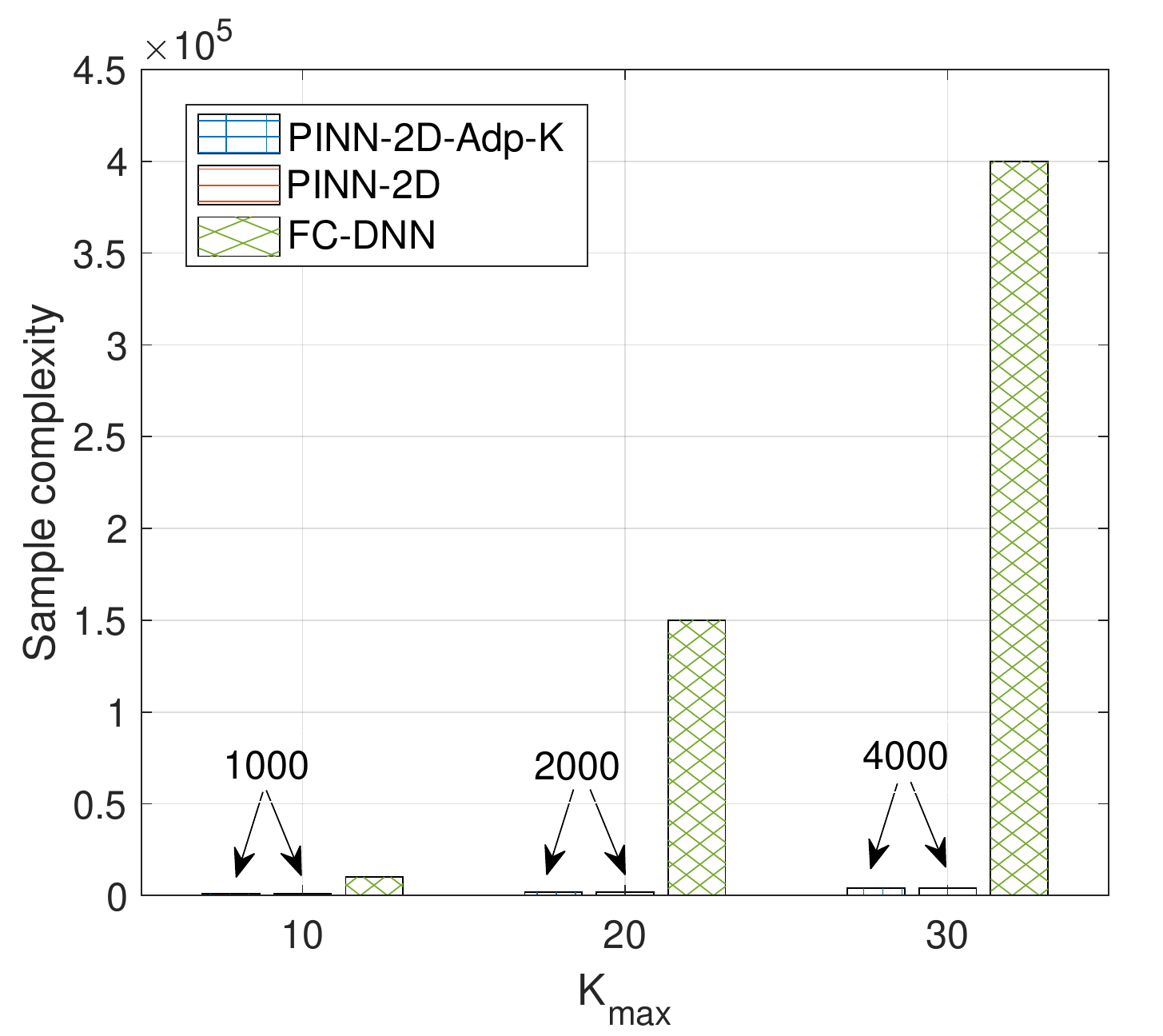}}
	\end{minipage}
	\begin{minipage}[t]{0.45\linewidth}	
		\subfigure[Computational complexity]{
			\includegraphics[width=\textwidth]{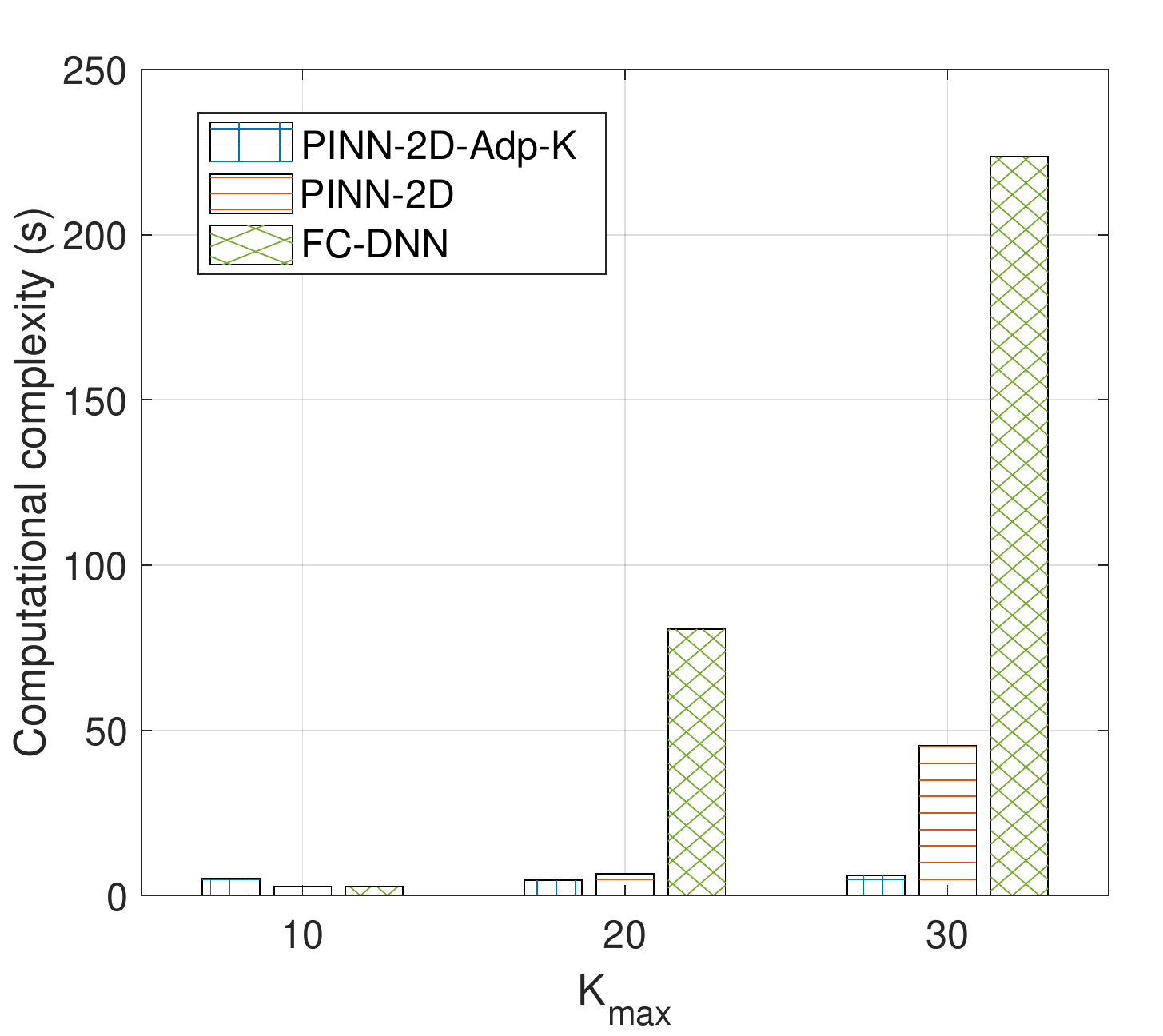}}
	\end{minipage}\vspace{-2mm}
	\caption{Training complexities of DNNs for learning power control in an interference network.}\label{fig: wmmse-cplxty}
\end{figure}\vspace{-0.1mm}

In Fig. \ref{fig: wmmse-cplxty}, we show the training complexity of the DNNs when $K_{\max}$ differs. As expected, both the complexities of training PINN-2D and PINN-2D-Adp-$K$ are much lower than ``FC-DNN'', and the complexity reductions grow with $K_{\max}$. When $K_{\max}=30$, the sample and computational complexities of training PINN-2D is respectively reduced by 99\% and 80\% from ``FC-DNN'', and the sample and computational complexities of training PINN-2D-Adp-$K$ is respectively reduced by 99\% and 97\%. Although the sample complexity of PINN-2D and PINN-2D-Adp-$K$ are almost the same, the complexity in generating labels for PINN-2D-Adp-$K$ is lower than PINN-2D. This is because most samples for training PINN-2D-Adp-$K$ are generated with $K<K_{\max}$, while all the samples for training PINN-2D are generated with $K=K_{\max}$.

\subsubsection{Permutation Invariance for Dataset Augmentation}
Generating labels is time-consuming, especially when $K_{\max}$ is large. This is because more samples are required for training (shown in Fig. \ref{fig: wmmse-cplxty} (a)), meanwhile generating each label costs more time to solve problem \eqref{P: max sum-rate}. In what follows, we show that the time consumed for generating labels can be reduced by dataset augmentation, i.e., generating more labels based on already obtained labels.

Specifically, by leveraging the permutation invariant relationship between ${\bf y}^*$ and ${\bf X}$, we know that for arbitrary permutation to ${\bf X}$, i.e., ${\bm \Lambda}^{\sf T}{\bf X}{\bm \Lambda}$, ${\bm \Lambda}^{\sf T}{\bf y}^*$ is the corresponding optimal solution. This suggests that we can generate a new sample $\{{\bm \Lambda}^{\sf T}{\bf X}{\bm \Lambda}, {\bm \Lambda}^{\sf T}{\bf y}^*\}$ based on an existed sample $\{{\bf X}, {\bf y}^*\}$. In this way of dataset augmentation, we can first generate a small number of training samples as in the setups in section \ref{sec:simu:IC:setup}, and then augment the dataset for more samples. For example, the possible permutations when $K_{\max}=30$ is $K_{\max}! \approx 2.65\times 10^{32}$, hence we can generate $2.65\times 10^{32}$ samples based on only a single sample!

In Table \ref{time-dataAug}, we compare the time consumption for generating training set with and without using dataset augmentation, when the trained ``FC-DNN''\footnote{The proposed PINNs cannot use the augmented samples for training, since the permutation invariance property has been used for constructing the architecture.} can achieve the same sum-rate on the training set. The legend ``generated samples'' means the samples generated as in section \ref{sec:simu:IC:setup}, and ``augmented samples'' means the samples augmented with the  permutation invariance.

\begin{table}[htb!]
	\centering
	\vspace{-2mm}
	\caption{Time consumed for generating samples, which is dominated by generating labels}\label{time-dataAug}
	\vspace{-2mm}
	\small
	\begin{tabular}{c|c|c|c|c}
		\hline\hline
		\multirow{2}{*}{$K_{\max}$} & \multicolumn{2}{c|}{\textbf{With dataset augmentation}} & \multicolumn{2}{c}{\textbf{Without dataset augmentation}} \\
		\cline{2-5}
		~ & \tabincell{c}{Number of generated samples +\\ Number  of augmented samples} & Time consumption & Number  of generated samples & Time consumption\\
		\hline
		\tabincell{c}{10} & 10~+~9,990 & 0.68 s &10,000 & 100 s\\
		\hline
		\tabincell{c}{20} & 10~+~149,990 & 10.27 s & 150,000 & 900 s\\
		\hline
		\tabincell{c}{30} & 10~+~399,990 & 30.6 s & 400,000 & 4000 s\\
		\hline\hline
	\end{tabular}
	\vspace{-5mm}
\end{table}

We can see from Table \ref{time-dataAug} that the number of training samples required by FC-DNN for achieving an expected performance is identical for the training set with and without dataset augmentation. However, the time complexity of generating samples with dataset augmentation can be reduced by about 99\% from that without dataset augmentation.

\section{Conclusions} \label{sec: conclusion}
In this paper, we constructed DNNs by sharing the weights among permutation invariant blocks and demonstrated how the proposed PINNs can adapt to the scales of wireless systems. We employed two case studies to illustrate how the PINNs can be applied, where the DNNs trained with and without supervision are used to learn the optimal solutions of predictive resource allocation and interference coordination, respectively. Simulation results showed that the numbers of model parameters of the PINNs are 1$/1000$ $\sim$ $1/10000$ of the fully-connected DNN when achieving the same performance, which leads to remarkably  reduced sample and computational complexity for training. We also found that the property of permutation invariance can be utilized for dataset augmentation such that the time consumed to generate labels for supervised learning can be reduced drastically. The proposed DNNs are applicable to a broad range of wireless tasks, thanks to the general knowledge incorporated.
\vspace{-1mm}
\begin{appendices}
\numberwithin{equation}{section}
\vspace{-1mm}
\section{Proof of proposition \ref{pp: 1}}\label{appendix: A}
\vspace{-1mm}
We first prove the necessity. Assume that the function $f({\bf x})$ is permutation invariant to ${\bf x}$. If the $k$th block ${\bf x}_k$ in ${\bf x}=[{\bf x}_1, \cdots, {\bf x}_K]$ is changed to another position in ${\bf x}$ while the permutation of other blocks in ${\bf x}$ remains unchanged, i.e.,
\begin{equation}
\textstyle\tilde{\bf x}=[\underbrace{{\bf x}_1,\cdots,{\bf x}_{k-1}}_{(a)},\underbrace{{\bf x}_{k+1},\cdots,{\bf x}_K}_{(b)}],
\end{equation}
 where ${\bf x}_k$ may be in the blocks in $(a)$ or $(b)$, then
 $\tilde{\bf y}_k={\bf y}_{k-1}$ if ${\bf x}_k$ is in $(a)$ and $\tilde{\bf y}_k={\bf y}_{k+1}$ if ${\bf x}_k$ is in $(b)$, hence $\tilde{\bf y}_k\neq {\bf y}_k$. This indicates that the $k$th output block should change with the $k$th input block ${\bf x}_k$. On the other hand, if the position of ${\bf x}_k$ remains unchanged while the positions of other blocks ${\bf x}$ arbitrarily change, i.e., $\tilde{\bf x}=[{\bf x}_{N_1},\cdots,{\bf x}_{N_k-1},{\bf x}_k,{\bf x}_{N_{k}+1},\cdots,{\bf x}_{N_K}]$, then $\tilde{\bf y}=[{\bf y}_{N_1},\cdots,{\bf y}_{N_k-1},{\bf y}_k,{\bf y}_{N_k+1},\cdots,{\bf y}_{N_K}]$, and $\tilde{\bf y}_k={\bf y}_k$. This means that ${\bf y}_k$ is not affected by the permutation of the input blocks other than ${\bf x}_k$. Therefore, the function should have the form in \eqref{eq: perm inva}.

We then prove the sufficiency. Assume that the function $f({\bf x})$  has the form in \eqref{eq: perm inva}. If ${\bf x}$ is changed to $\tilde{\bf x}=[{\bf x}_{N_1},\cdots, {\bf x}_{N_K}]$, then the $k$th block of $\tilde{\bf y}$ is $\tilde{\bf y}_k = \eta(\psi({\bf x}_{N_k}),{\cal F}_{n=1,n\neq N_k}^K \phi({\bf x}_n))={\bf y}_{N_k}$. Hence, the output corresponding to $\tilde{\bf x}$ is $\tilde{\bf y}=[\tilde{\bf y}_1,\cdots,\tilde{\bf y}_K]=[{\bf y}_{N_1},\cdots,{\bf y}_{N_K}]$. According to Definition \ref{def: 1}, the function in \eqref{eq: perm inva} is permutation invariant to ${\bf x}$.
\vspace{-2mm}

\end{appendices}
\vspace{-1mm}
	\bibliography{IEEEabrv,GJ1}
	\vspace{-1mm}
\end{document}